\documentclass[11pt]{article}

\usepackage[final]{acl}

\usepackage{times}
\usepackage{latexsym}
\usepackage{pifont}

\usepackage[T1]{fontenc}

\usepackage[utf8]{inputenc}

\usepackage{microtype}

\usepackage{inconsolata}

\usepackage{graphicx}

\usepackage{booktabs}
\usepackage{longtable}
\usepackage{multirow}
\usepackage{bigstrut}
\usepackage{makecell}
\usepackage[table]{xcolor}
\usepackage{array}
\usepackage{microtype}
\usepackage{amsmath}
\usepackage{hyperref}
\usepackage{tcolorbox}
\tcbuselibrary{skins}
\renewcommand{\arraystretch}{1.15}
\newcolumntype{L}[1]{>{\raggedright\arraybackslash}p{#1}}

\usepackage{enumitem}

\setlist[itemize]{
  parsep=0pt,
  leftmargin=1em   
}

\definecolor{nodecontext}{HTML}{E0E0E0}
\definecolor{nodeplanning}{HTML}{FFADAD}
\definecolor{nodefact}{HTML}{FFD6A5}
\definecolor{nodereasoning}{HTML}{FDFFB6}
\definecolor{noderestatement}{HTML}{CAFFBF}
\definecolor{nodeassumption}{HTML}{BDFBDF}
\definecolor{nodeexample}{HTML}{9BF6FF}
\definecolor{nodereflection}{HTML}{A0C4FF}
\definecolor{nodeconclusion}{HTML}{C3B1E1}
\definecolor{edgereasoninfer}{HTML}{E0E0E0}
\definecolor{edgereasonexecute}{HTML}{D8D8D8}
\definecolor{edgereasonrestate}{HTML}{D0D0D0}
\definecolor{edgereasonelaboratefact}{HTML}{C8C8C8}
\definecolor{edgereasonexemplify}{HTML}{C0C0C0}
\definecolor{edgeplanproceed}{HTML}{FFCDCD}
\definecolor{edgeplanverify}{HTML}{FFC4C4}
\definecolor{edgeplandecompose}{HTML}{FFB9B9}
\definecolor{edgeplanbacktrack}{HTML}{FFADAD}
\definecolor{edgereflectpositive}{HTML}{D0E4FF}
\definecolor{edgereflectuncertain}{HTML}{C8DBFF}
\definecolor{edgereflectnegative}{HTML}{C0D4FF}
\definecolor{edgevalidatesupport}{HTML}{FFE5C5}
\definecolor{edgevalidateattack}{HTML}{FFCD8E}

\newcommand{\nodecontext}[0]{\colorbox{nodecontext}{\footnotesize\texttt{Context}}}
\newcommand{\nodeplanning}[0]{\colorbox{nodeplanning}{\footnotesize\texttt{Planning}}}
\newcommand{\nodefact}[0]{\colorbox{nodefact}{\footnotesize\texttt{Fact}}}
\newcommand{\nodereasoning}[0]{\colorbox{nodereasoning}{\footnotesize\texttt{Reasoning}}}
\newcommand{\noderestatement}[0]{\colorbox{noderestatement}{\footnotesize\texttt{Restatement}}}
\newcommand{\nodeassumption}[0]{\colorbox{nodeassumption}{\footnotesize\texttt{Assumption}}}
\newcommand{\nodeexample}[0]{\colorbox{nodeexample}{\footnotesize\texttt{Example}}}
\newcommand{\nodereflection}[0]{\colorbox{nodereflection}{\footnotesize\texttt{Reflection}}}
\newcommand{\nodeconclusion}[0]{\colorbox{nodeconclusion}{\footnotesize\texttt{Conclusion}}}

\newcommand{\edgereasoninfer}[0]{\colorbox{edgereasoninfer}{\footnotesize\texttt{infer}}}
\newcommand{\edgereasonexecute}[0]{\colorbox{edgereasonexecute}{\footnotesize\texttt{execute}}}

\newcommand{\edgereasonelaboratefact}[0]{\colorbox{edgereasonelaboratefact}{\footnotesize\texttt{elaborate-fact}}}
\newcommand{\edgereasonexemplify}[0]{\colorbox{edgereasonexemplify}{\footnotesize\texttt{exemplify}}}
\newcommand{\edgeplanproceed}[0]{\colorbox{edgeplanproceed}{\footnotesize\texttt{proceed}}}
\newcommand{\edgeplanverify}[0]{\colorbox{edgeplanverify}{\footnotesize\texttt{verify}}}
\newcommand{\edgeplandecompose}[0]{\colorbox{edgeplandecompose}{\footnotesize\texttt{decompose}}}
\newcommand{\edgeplanbacktrack}[0]{\colorbox{edgeplanbacktrack}{\footnotesize\texttt{backtrack}}}
\newcommand{\edgereflectpositive}[0]{\colorbox{edgereflectpositive}{\footnotesize\texttt{positive}}}
\newcommand{\edgereflectnegative}[0]{\colorbox{edgereflectnegative}{\footnotesize\texttt{negative}}}
\newcommand{\edgereflectuncertain}[0]{\colorbox{edgereflectuncertain}{\footnotesize\texttt{uncertain}}}
\newcommand{\edgevalidatesupport}[0]{\colorbox{edgevalidatesupport}{\footnotesize\texttt{support}}}
\newcommand{\edgevalidateattack}[0]{\colorbox{edgevalidateattack}{\footnotesize\texttt{attack}}}

\tcbset{
  findingsbox/.style={
    colback=blue!8!white,
    colframe=blue!50!black,
    coltitle=blue!30!black,
    arc=4pt,
    boxrule=0.8pt,
    left=8pt, right=8pt,
    top=4pt, bottom=6pt,
  }
}

\tcbset{
  promptbox/.style={
    colback=gray!8!white,
    colframe=gray!50!black,
    coltitle=gray!30!black,
    arc=4pt,
    boxrule=0.8pt,
    left=8pt, right=8pt,
    top=4pt, bottom=6pt,
  }
}

\definecolor{IlliniOrange}{RGB}{232, 74, 39}

\title{ReasoningFlow: Discourse Structures for Understanding \\ LLM Reasoning Traces}

\author{\textbf{Jinu Lee}, \textbf{Shivam Agarwal}, \textbf{Amruta Parulekar}, \textbf{Siddarth Madala}, \\ \textbf{Dilek Hakkani-T\"ur}, \textbf{Julia Hockenmaier} \\
  The Grainger College of Engineering \\
  University of Illinois Urbana-Champaign \\
  \texttt{\{jinulee2, shivam2, amp20, smadala2, dilek, juliahmr\}@illinois.edu} \\}

\begin{document}
\maketitle
\begin{abstract}

Large reasoning models (LRMs) produce reasoning traces with non-linear structures, such as backtracking and self-correction, that complicate the evaluation and monitoring of the reasoning process.
We introduce \textbf{ReasoningFlow}, a framework that captures the discourse structures of LRM reasoning traces into fine-grained directed acyclic graphs (DAGs).
We develop and validate our annotation schema through careful manual annotation of 31 traces (2.1k steps), achieving high inter-annotator agreement, then scale to automatic annotation of 1,260 traces (247.7k steps) spanning three tasks (math, science, argumentation) and five models (Qwen2.5-32B-Inst, QwQ-32B, DeepSeek-V3, DeepSeek-R1, GPT-oss-120B).
By analyzing ReasoningFlow graphs, we find:
(1) LRMs exhibit structurally similar traces, despite being trained from different base models and potentially non-overlapping post-training data.  (2) ReasoningFlow reveals diverse fine-grained reasoning behaviors (e.g., local verification, self-reflection, and assumptions) that can be used for better reasoning trace monitorability. (3) In LRMs, most of the erroneous steps are not used to derive final answers. (4) Mechanistic causal dependencies between steps do not reflect the language-level discourse structure. 
We release the dataset and code in: \href{https://github.com/jinulee-v/reasoningflow}{Homepage}.

\end{abstract}

\section{Introduction}



Large Reasoning Models (LRMs; e.g., DeepSeek-R1 \citep{guo_deepseek-r1_2025}) generate extended reasoning traces with non-linear reasoning behaviors, such as verification, self-reflection, and backtracking \citep{gandhi_cognitive_2025}. This non-linearity complicates both correctness evaluation and faithfulness monitoring. For instance, stepwise evaluation \citep{lightman_lets_2024} may flag an erroneous step, yet the trace as a whole may still be correct if the self-verification overrides the previous error.

Recent attempts to understand the non-linear structure of LRM traces either lack expressive relation labels or only annotate inter-paragraph structures \citep{bogdan_thought_2025, jiang_what_2025, marjanovic_deepseek-r1_2026}, which are too coarse for annotating fine-grained reasoning behaviors. On the other hand, discourse structure annotations for human text \citep{carlson_building_2001, stab_parsing_2017} fail to capture the relations and structures emerging in goal-oriented reasoning traces.


\begin{figure}
    \centering
    \includegraphics[width=0.95\linewidth]{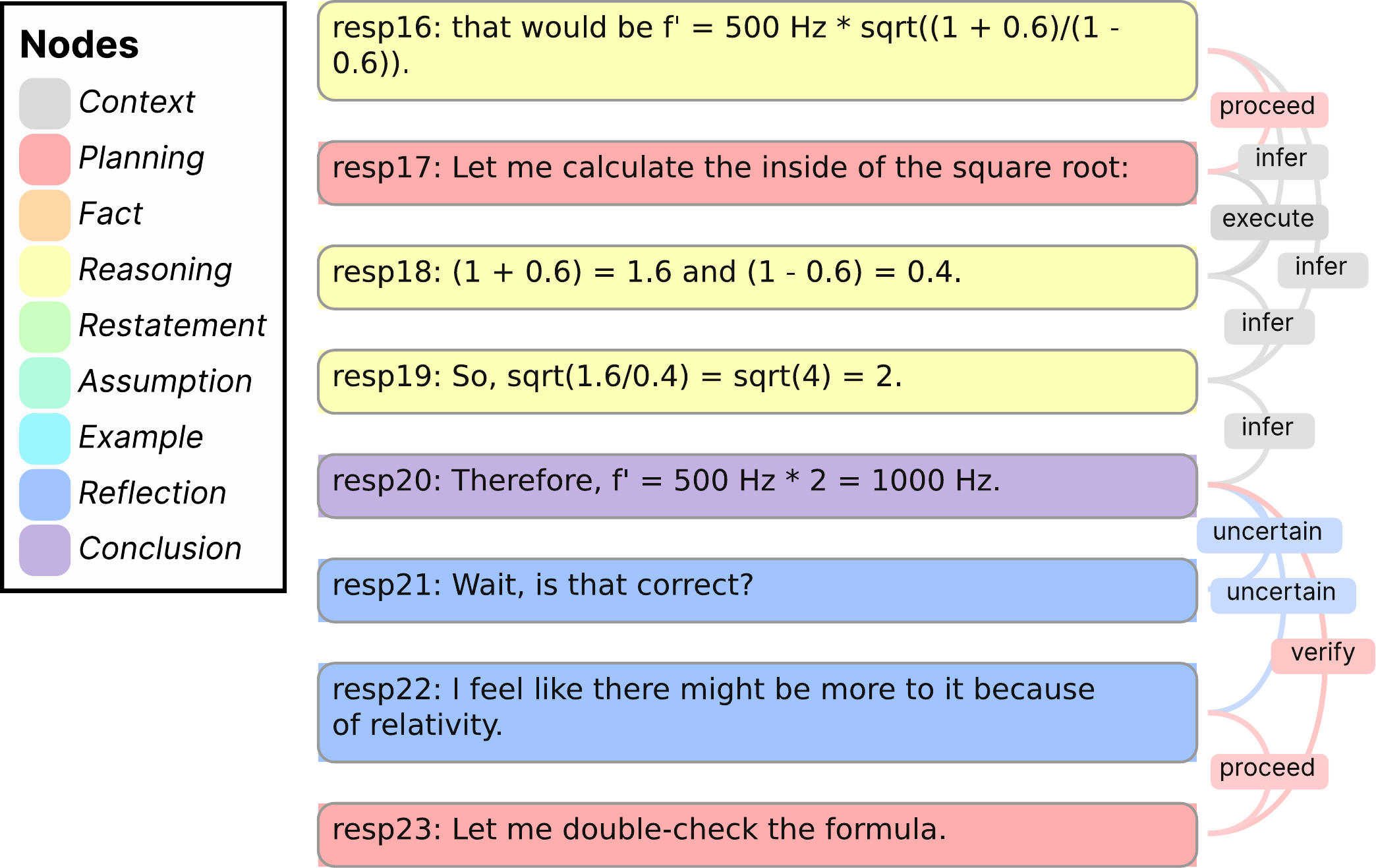}
    \caption{Example of a ReasoningFlow graph. ReasoningFlow segments the reasoning trace into \textbf{nodes}, and annotates the relations between the nodes as \textbf{edges}. This example shows deductive reasoning (\texttt{resp16-20}) and self-reflection/verification (\texttt{resp20-23}) behaviors.}
    \label{fig:main-example}
\end{figure}

\begin{table*}[h]
\centering
\footnotesize
\begin{tabular}{l|c|c|c|c|c|c}
\toprule
\textbf{Schema} & \textbf{LRM?} & \textbf{\# Node} & \textbf{\# Edge} & \textbf{Granularity} & \textbf{Structure} & \textbf{IAA?} \\
\midrule
PARC \citep{mukherjee_premise-augmented_2025} & X & 1 & 1 & Paragraph & \textbf{DAG} & O \\
Thought Anchors \citep{bogdan_thought_2025} & O & \textbf{8} & 1 & Sent. & \textbf{DAG} & O \\
R1-Thoughtology \citep{marjanovic_deepseek-r1_2026} & O & 4 & - & Paragraph & Linear & X \\
LCoT2Tree \citep{jiang_what_2025} & O & 1 & 4 & Paragraph & Tree & X \\
ReJump \citep{zeng_rejump_2025} & O & 1 & 3 & Paragraph & Tree & X \\
\textbf{ReasoningFlow (this work)} & O & \textbf{8} & \textbf{14} & \textbf{Sub-sent.} & \textbf{DAG} & \textbf{O} \\
\bottomrule
\end{tabular}
\caption{Comparison of structure annotation schema for LLM reasoning traces. Each column corresponds to: (LRM?) whether it is designed to accommodate LRMs' long traces, (\# Node) number of node labels, (\# Edge) number of edge labels, (Granularity) granularity of nodes, (Structure) the generated graph structure, and (IAA?) whether multiple human annotators verified the schema. At the time of writing, ReasoningFlow is the only work to annotate fine-grained nodes and edges, and validate with inter-annotator agreement analysis.}
\label{tab:related-work-comparison}
\end{table*}

We develop \textbf{ReasoningFlow}, a framework for annotating fine-grained discourse structures of reasoning traces. ReasoningFlow converts reasoning traces into a directed acyclic graph with 8 node types and 14 edge types. We release 31 manually annotated and cross-verified reasoning traces (2.1k steps), along with 1,260 automatically annotated traces (247.7k steps) generated by five models (Qwen2.5-32B-Inst, QwQ-32B, DeepSeek-V3, DeepSeek-R1, GPT-oss-120B) across math, science, and argumentation tasks.

ReasoningFlow can be used to improve the monitorability and faithfulness of reasoning traces. Using ReasoningFlow, we find:

\begin{itemize}
    \item LRMs across different families and sizes demonstrate similar reasoning trace structures. 
    \item ReasoningFlow can identify fine-grained reasoning behaviors like \textit{local verification, self-reflection, and assumptions}, enabling a new dimension for monitoring reasoning traces.
    \item Most erroneous steps in LRMs do not lead to incorrect final answers, explaining why error detection does not directly transfer to better performance in LRMs.
    \item Mechanistically measured step-to-step causal dependencies \citep{bogdan_thought_2025} do not reflect text-level discourse relations. 
\end{itemize}

\section{Related Works}

\subsection{Reasoning trace structures}

Prior to LRMs, reasoning traces were commonly viewed as entailment graphs, which anchor each step to its logical premises \citep{ling_deductive_2023, mukherjee_premise-augmented_2025}. Yet, these graphs only show logical entailments; therefore, they cannot capture the diverse reasoning patterns exhibited by LRMs, including planning and verification.

Early efforts to analyze LRM traces focus on modeling verification. \citet{marjanovic_deepseek-r1_2026} treats an LRM trace as an initial solution followed by iterative verification attempts, while \citet{jiang_what_2025, zeng_rejump_2025} models traces as \textit{trees of paragraphs} \citep{yao_tree_2023} with verification and backtracking hyperedges. However, both approaches are too coarse to characterize the full spectrum of reasoning behaviors, which can occur down to the sub-sentence level (Section \ref{sec:behaviors}).

A more fundamental limitation cuts across all of this work: none validates its annotation scheme inter-annotator agreement, leaving open the question of whether the proposed frameworks are consistently interpretable. Given the linguistic complexity of LRM reasoning traces, there is a strong need for a framework that is both expressive enough to capture diverse reasoning patterns and reliable enough to be applied consistently by human annotators. Table \ref{tab:related-work-comparison} compares related works on LLM reasoning trace structure annotation.


\subsection{Discourse/Argumentation structures}

Discourse and argumentation parsing frameworks have been widely applied to capture the semantic structures of long texts.

\textbf{Discourse structures.} Rhetorical Structure Theory (RST) builds hierarchical discourse structures with directed relations between two clauses \citep{mann_rhetorical_1988}. RST defines over 20 relation types between clauses (e.g., elaboration, cause, condition), covering diverse rhetorical intent of the author \citep{carlson_building_2001}.

\textbf{Argumentation structures} capture the argumentational role of a text span (major claim, minor claim, premise) and the semantic relation between two spans (\textit{Does this span support or attack the corresponding claim?}) \citep{stab_parsing_2017}. The resulting document structure is typically represented as a tree, in which atomic premises connect recursively up to a major claim.

\begin{table*}[t]
\centering
\footnotesize
\setlength{\tabcolsep}{2pt}
\setlength{\fboxsep}{2pt}
\renewcommand{\arraystretch}{1.15}
\begin{tabular}{@{}p{3cm}p{12.4cm}@{}}
\toprule
\textbf{Label} & \textbf{Definition} \\
\midrule
\multicolumn{2}{@{}l}{\textit{Node labels}} \\
\midrule
\nodecontext & User-provided text: problem statement, retrieved documents, tool responses. \\
\nodeplanning & States what to do next; sets a global direction or a local goal for the following nodes. \\
\nodefact & Parametric knowledge (laws, definitions, constants) independent of the context and of prior nodes. \\
\nodereasoning & Deductive, inductive, or abductive inference that derives new content from previous nodes. \\
\noderestatement & Copies or paraphrases the context or an earlier node, adding no new information. \\
\nodeassumption & A premise marked as not necessarily true, scoping the nodes that depend on it. \\
\nodeexample & A concrete instance illustrating a general concept introduced earlier. \\
\nodereflection & Subjective judgment or emotion about earlier nodes. \\
\nodeconclusion & An intermediate or final answer to the input, asserted by the model. \\
\midrule
\multicolumn{2}{@{}l}{\textit{Edge labels}} \\
\midrule
\colorbox{edgereasoninfer}{\strut\texttt{infer}} & Premise $\rightarrow$ result (logical deduction, calculation, or inductive hypothesis) \\
\colorbox{edgereasonexecute}{\strut\texttt{execute}} & \nodeplanning $\rightarrow$ the step that implements the plan. \\
\colorbox{edgereasonrestate}{\strut\texttt{restate}} & Original node $\rightarrow$ the node citing/restating the original node. \\
\colorbox{edgereasonelaboratefact}{\strut\texttt{elaborate-fact}} & General \nodefact $\rightarrow$ a more specific \nodefact~detailing it. \\
\colorbox{edgereasonexemplify}{\strut\texttt{exemplify}} & Concept $\rightarrow$ an \nodeexample~illustrating it. \\
\colorbox{edgeplanproceed}{\strut\texttt{proceed}} & Content that motivates a plan $\rightarrow$ \nodeplanning. \\
\colorbox{edgeplanverify}{\strut\texttt{verify}} & Original node $\rightarrow$ \nodeplanning~that initiates its verification. \\
\colorbox{edgeplandecompose}{\strut\texttt{decompose}} & Coarse \nodeplanning~node $\rightarrow$ \nodeplanning~ with finer subplan or sub-case. \\
\colorbox{edgeplanbacktrack}{\strut\texttt{backtrack}} & \nodeplanning~ $\rightarrow$ alternative \nodeplanning~ (backtracking). \\
\colorbox{edgereflectpositive}{\strut\texttt{positive}} & Node $\rightarrow$ \nodereflection~affirming it. \\
\colorbox{edgereflectnegative}{\strut\texttt{negative}} & Node $\rightarrow$ \nodereflection~ denying it or showing doubt. \\
\colorbox{edgereflectuncertain}{\strut\texttt{uncertain}} & Node $\rightarrow$ \nodereflection~ expressing uncertainty or confusion. \\
\colorbox{edgevalidatesupport}{\strut\texttt{support}} & Node $\rightarrow$ independently derived node asserting the same proposition. \\
\colorbox{edgevalidateattack}{\strut\texttt{attack}} & Node $\rightarrow$ node that contradicts or explicitly corrects it. \\
\bottomrule
\end{tabular}
\caption{Summary of 8 node labels and 14 edge labels of ReasoningFlow. Detailed definitions and examples can be found in Tables \ref{tab:node-labels} and \ref{tab:edge-labels}.}
\label{tab:labels}
\end{table*}

However, both approaches are not fully compatible with LLM reasoning traces for two reasons. First, existing schemas are not designed to accommodate reasoning trace-specific phenomena like \textit{verification}, which rarely appear in well-organized news or argumentative texts. Second, autoregressively generated LRM traces exhibit only left-to-right causal dependencies as in improvised human speech \citep{kempen_incremental_1987}, where organized texts frequently exhibit \textit{backward} dependencies. Together, these underscore the need for an annotation schema designed specifically for reasoning traces.

Appendix \ref{sec:appendix-literature-review} includes a detailed comparison between ReasoningFlow and related works across LLM reasoning, computational linguistics, formal logic, and cognitive science.

\section{ReasoningFlow schema}

We introduce ReasoningFlow, a framework for annotating fine-grained semantic structures of reasoning traces.

We adopt a \textit{directed acyclic graph} (DAG) structure with edges always flowing from earlier steps to later steps, resembling the left-to-right information flow in autoregressive LLMs \citep{ling_deductive_2023, bogdan_thought_2025}. Compared to the projective Rhetorical Structure Theory-based trees \citep{carlson_building_2001} and single-root argumentation trees \citep{stab_parsing_2017}, DAG provides both structural flexibility (e.g., crossing edges, one step having multiple successors) and a straightforward automatic annotation algorithm (Section \ref{sec:automatic-annotation}).

\textbf{Nodes.} Nodes are contiguous, non-overlapping snippets that contain elementary reasoning steps. We primarily treat each sentence as a single node, but we divide a sentence into multiple nodes when two clauses are connected with distinct functional roles. For instance, if a step reads \textit{"Therefore, $x$ should be 17, but I should double-check."}, it is more natural to assign different roles to the first (calculating the answer) and the second half (planning the verification).

We define 8 node types based on their functional roles. The three core types are \nodereasoning, \nodeplanning, and \nodereflection. \nodereasoning~nodes contain main building blocks like deduction and calculation, \nodeplanning~nodes introduce the content of upcoming nodes, and \nodereflection~nodes evaluate the correctness or express certainty on previous nodes. In addition, we define five special cases of \nodereasoning~nodes, namely \nodefact, \noderestatement, \nodeassumption, \nodeexample, and \nodeconclusion, that indicate special roles for downstream applications. Table \ref{tab:node-labels} lists full definitions and examples for all node labels.

\textbf{Edges}. The next step is to annotate semantic relations between nodes as directed edges. All edges are constrained to flow left-to-right, uniquely connecting earlier nodes to later nodes in the sequence.

We define 14 edge labels of four major categories: Reason, Plan, Reflect, and Validate. Reason-related edges describe how the current step is derived from previous premises, plans, and facts. Plan-related edges show the intent of the \nodeplanning~node, e.g., to proceed to a next step, or to verify the previous findings. Reflect-related edges show what \nodereflection~nodes evaluate, along with the sentiment. Finally, Validate-related edges compare the propositional equivalence between distant nodes, deciding whether the following nodes \edgevalidatesupport~or \edgevalidateattack~the previous statement. Detailed definitions and examples for edge labels can be found in Table \ref{tab:edge-labels}.


\section{Dataset construction}
\label{sec:dataset-construction}

\subsection{Manual annotation}
\label{sec:manual-annotation}

To validate the ReasoningFlow schema's clarity, we performed manual annotation with inter-annotator agreement evaluation. The manually annotated portion consists of 31 math, physics, and chemistry questions selected from NuminaMath \citep{zeng_numina_2025} and STILL-2 \citep{min_imitate_2024}, and traces were generated by QwQ-32B-Preview \citep{qwen_team_qwq_2024} with temperature 0.

Four of the authors participated in manual annotation, where two annotators were assigned to each trace. All annotators were provided with the same segmentation manually done by one of the authors. Two annotators individually completed the annotation without interacting, which was used for inter-annotator agreement analysis below. Then, annotators discussed the discrepancies in the annotation, which were incorporated in the guidelines, subtype definitions, and few-shot examples used for LLM-based annotation.

We measure two types of inter-annotator agreements, namely Node Classification (NC) and Edge Detection/Classification (EDC). NC measures whether both annotators selected the same node label (number of categories $k=8$); EDC measures whether two annotators agree on whether the two nodes are connected or not, and the edge label if connected ($k=15$; \texttt{no-edge} and 14 edge labels). The results show that annotators agree significantly on NC and EDC with Krippendorff's $\alpha>0.8$ (Table \ref{tab:inter-annotator-agreement}), which is considered highly reliable \citep{krippendorff_reliability_2004}. This level of agreement indicates that the ReasoningFlow categories are well-defined and consistently interpretable across annotators.

For node classification, the most common disagreement is whether a node should be classified as \nodereasoning~or others. For instance, “\textit{Given that Z is the midpoint, we can use the midpoint formula to find the coordinates of X.}” was annotated as \nodereasoning~and \nodeplanning~between different annotators. Edge classification disagreements commonly stemmed from node classification disagreements (e.g., \edgereasoninfer~or \edgereasonexecute~edge labels apply when the previous node is \nodereasoning~or \nodeplanning, respectively).

\begin{table}[t]
\centering
\footnotesize
\begin{tabular}{lcc}
\hline
\textbf{Metric} & \textbf{Krippendorff's } $\alpha$ & \textbf{N} \\
\hline
NC      & 0.8851 & 1,657 \\
EDC     & 0.9193 & 122,630 \\
\hline
\end{tabular}
\caption{Inter-annotator agreement (Krippendorff's $\alpha$) measured between four human annotators. Each example was assigned to two annotators. High $\alpha>0.8$ validates the annotation schema of ReasoningFlow.}
\label{tab:inter-annotator-agreement}
\end{table}

\subsection{Automatic annotation}
\label{sec:automatic-annotation}

We perform large-scale annotation of ReasoningFlow using an LLM-powered automatic annotation pipeline.

\subsubsection{Base trace generation}

We choose three representative datasets: AIME 2024 \citep{mathematical_association_of_america_american_2024}, GPQA-Diamond \citep{rein_gpqa_2023}, and ArgKP \citep{bar-haim_arguments_2020}. AIME 2024 contains 30 competition-level math problems. GPQA-Diamond is a benchmark for scientific knowledge and reasoning, including 198 problems across physics, chemistry, and biology. Finally, ArgKP is an argumentation benchmark with 24 debatable statements (e.g., \textit{We should prohibit flag burning.}), where the objective is to choose a stance (agree/disagree) and provide reasons.

A total of five models were used to collect the reasoning trace. For LRMs, we choose three representative models: \texttt{DeepSeek-R1} \citep[][671B]{guo_deepseek-r1_2025}, \texttt{QwQ-32B} \citep{qwen_team_qwq_2024}, and \texttt{GPT-oss-120B} \citep[][Reasoning effort: Medium]{openai_gpt-oss-120b_2025}. We also include two non-reasoning models: \texttt{DeepSeek-V3} \citep[][671B]{deepseek-ai_deepseek-v3_2024} and \texttt{Qwen2.5-32B-Instruct} \citep{yang_qwen25_2024}. We use greedy decoding (temperature 0) for all models.

As a result, we obtain 1,260 reasoning traces across five models and three datasets.

\subsubsection{Annotation pipeline}
\label{subsec:annotation-pipeline}

The annotation pipeline consists of three stages: node segmentation, node classification, and edge detection/classification.

During node segmentation, we instruct LLMs to segment raw reasoning traces into nodes, providing definitions and examples of where to segment and where not to. In the second stage (node classification), we prompt LLMs with node label definitions and representative examples.

Finally, in the edge detection/classification stage, we find all edges and their relation labels in a single pass. Since the unconstrained DAG structure of ReasoningFlow admits up to $O(N^2)$ edges, we ask LLMs to identify incoming edges for a single node for each inference to reduce output length and improve annotation performance. Details are presented in Appendix \ref{subsec:appendix-automatic-annotation-implementation-details}.

We used Gemini-3.1-Flash for node annotation and Gemini-3-Pro for edge annotation, as they achieve the best F1-scores on the manually annotated set with NC (0.865) and EDC (0.646), respectively; see Appendix \ref{subsec:appendix-model-performance-comparison} for scores from different models. Furthermore, the authors manually verified 30 of the automatic annotations by editing LLM predictions following \citet{mukherjee_premise-augmented_2025, zeng_rejump_2025}, achieving F1 scores of NC 0.909 and EDC 0.869 (Appendix \ref{subsec:appendix-manual-verification}).

\subsection{Node quality annotation}
\label{subsec:node-quality}

We annotate the \textit{quality} of ReasoningFlow nodes for further analyses (Sections \ref{sec:behaviors}-\ref{sec:validity-evaluation}).

For reasoning tasks (AIME, GPQA), we define the quality of a node as \textit{validity}, or logical correctness \citep{lightman_lets_2024, lee_evaluating_2025}. For each \nodereasoning~node, we employ LLM-as-a-judge (Gemini-3-Flash) to annotate whether the step is logically correct. We provide connected preceding nodes to the judge instead of the full context to reduce input length, following existing works \citep{ling_deductive_2023, mukherjee_premise-augmented_2025}.

For the argumentation task (ArgKP), we leverage the AQR dataset \citep{gretz_large-scale_2020}, which includes 30k crowdsourced arguments corresponding to ArgKP dataset's topics and human-annotated quality scores (0-1) for each argument. Specifically, we identify arguments in AQR that are equivalent to each \nodefact~and \nodereasoning~nodes using Gemini-3-Flash, and use the average score of matching arguments as the node quality.

Appendix \ref{sec:appendix-node-quality} includes detailed methods and manual verification results.

\section{ResaoningFlow statistics}
\label{sec:statistics}

\subsection{Nodes/edges count}

\begin{figure}
    \centering
    \includegraphics[width=\linewidth]{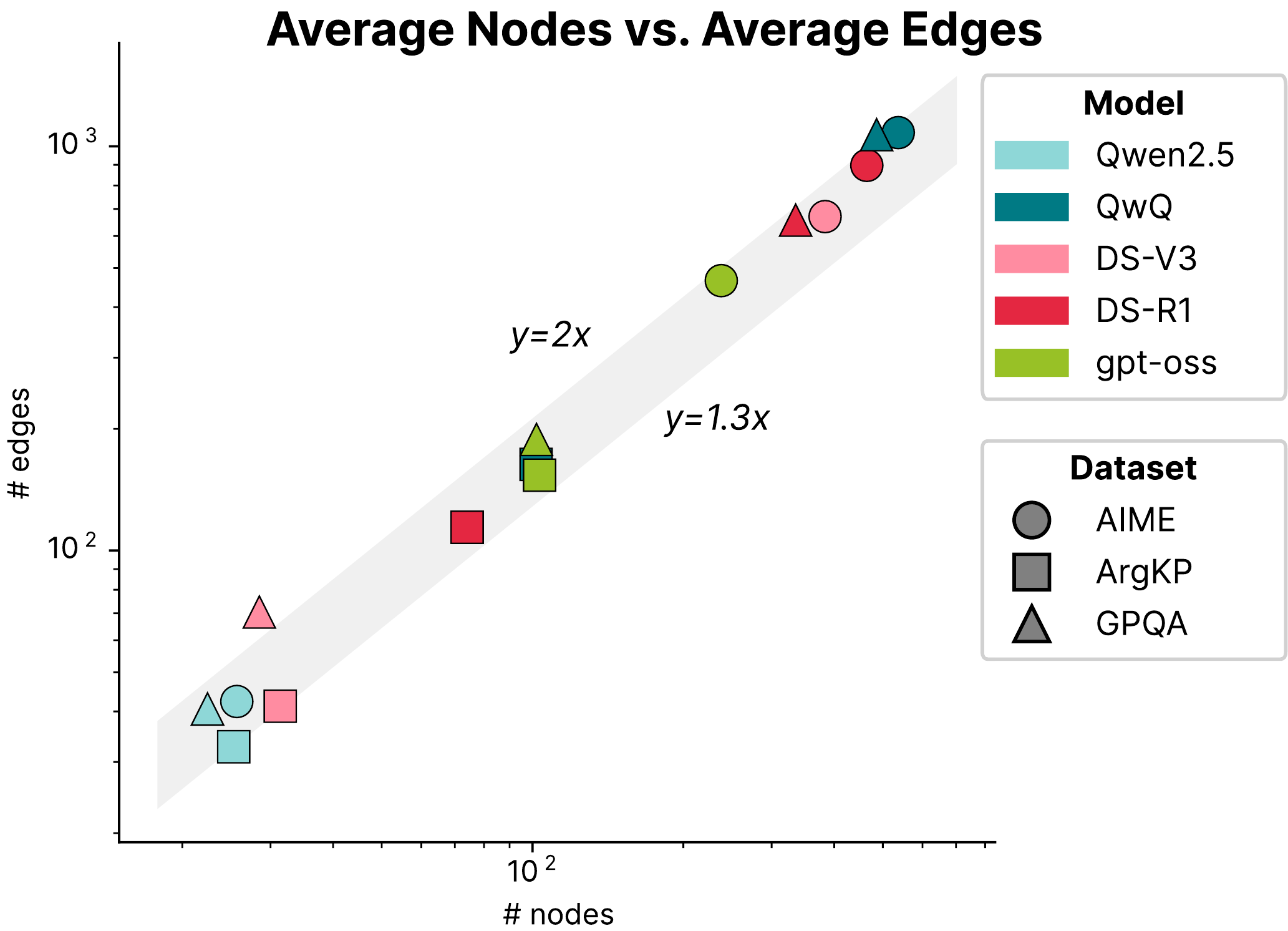}
    \caption{Average number of nodes and edges per (model, dataset). While the graph size varies, the average degree (edge/node) remains in 1.3-2.0 (gray region).}
    \label{fig:node-edge-degree}
\end{figure}

\begin{figure*}
    \centering
    \includegraphics[width=\linewidth]{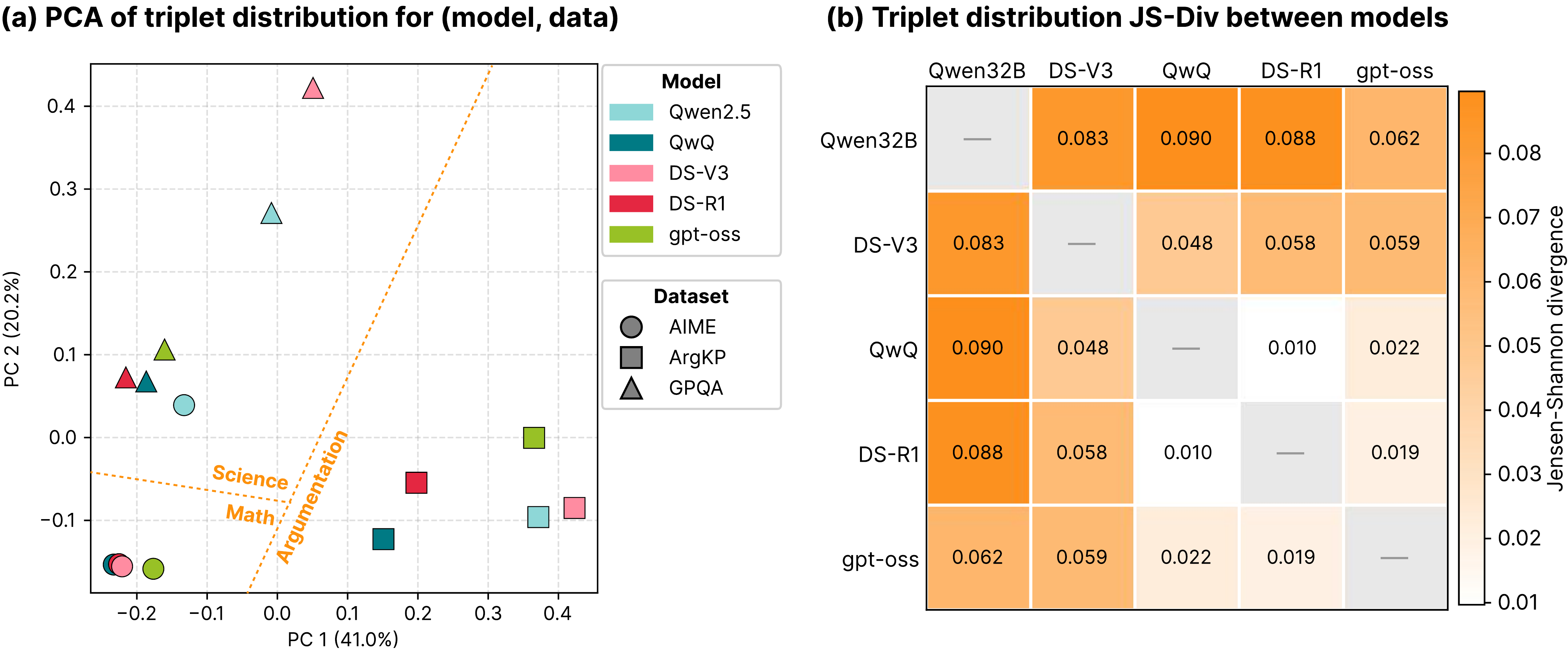}
    \caption{(a) Hellinger Principal Component Analysis (PCA) plot \citep{lebret_word_2014} of triplet probability distribution, showing clusters of datasets over clusters of models. (b) Jensen-Shannon Divergence of triplet distributions between models.}
    \label{fig:triplet-distribution}
\end{figure*}

The size of the graph varies considerably across models and datasets; QwQ generates an average of 455.6 nodes for AIME, while Qwen2.5-32B generates only 29.2 nodes for ArgKP. In contrast, the average degree (number of incoming edges per node) remains relatively stable across configurations, consistently falling between 1.3 and 2.0 (Figure \ref{fig:node-edge-degree}). This is consistent with prior findings in non-reasoning models, where each step typically draws on only a small number of premises \citep{ling_deductive_2023, mukherjee_premise-augmented_2025}.





\subsection{Comparison between models/domains}

\begin{tcolorbox}[findingsbox, title={\textcolor{white}{Finding 1}}]
  Different LRMs exhibit similar ReasoningFlow structures (triplet distributions).
\end{tcolorbox}

To investigate how reasoning structures vary across models and domains, we compare the distribution of (node label, edge label, node label) \textbf{triplets} extracted from ReasoningFlow graphs. Each triplet type relates to a specific reasoning behavior; e.g., \nodereasoning--\edgeplanverify$\rightarrow$\nodeplanning~triplets indicate verification, while \nodeassumption--\edgevalidateattack$\rightarrow$\nodereasoning~corresponds to proof-by-contradiction. For each (model, domain) pair, we compute the distribution of these triplets.

We apply Hellinger Principal Component Analysis (PCA) for explainable clustering of the triplet distribution of (model, dataset) pairs \citep{ding_emphk-means_2004, lebret_word_2014} (Figure \ref{fig:triplet-distribution}(a)). The main clusters form based on the domain rather than the generated model, indicating that different reasoning tasks elicit unique reasoning structures.

Principal components further reveal important features distinguishing the datasets. Argumentation traces include several subarguments listed in a sequence, which is realized by frequent \nodeplanning--\edgeplanproceed$\rightarrow$\nodeplanning. Math reasoning (AIME) involves more deductive reasoning steps \nodereasoning--\edgereasoninfer$\rightarrow$\nodereasoning, while science reasoning (GPQA) triggers more solution-level verification \nodeconclusion--\edgevalidatesupport$\rightarrow$\nodeconclusion. Refer to Appendix \ref{sec:appendix-stats} for details on PCA analysis.

Across models, reasoning model (QwQ, DS-R1, GPT-oss) traces are more structurally similar to one another than their base model counterparts, particularly on reasoning-heavy tasks. Figure \ref{fig:triplet-distribution}(b) displays the Jensen-Shannon divergence \citep[JS-Div;][]{fuglede_jensen-shannon_2004} between the triplet distribution of different models, all datasets averaged. While Qwen2.5-32B and DeepSeek-V3 exhibit substantial structural differences (JS-Div$=0.083$), their corresponding reasoning model checkpoints (QwQ and DeepSeek-R1) are far more similar (JS-Div$=0.010$). This observation suggests that different reasoning models, despite being trained with different base models and data, exhibit structural similarity in their reasoning traces.

\begin{figure*}
    \centering
    \includegraphics[width=\linewidth]{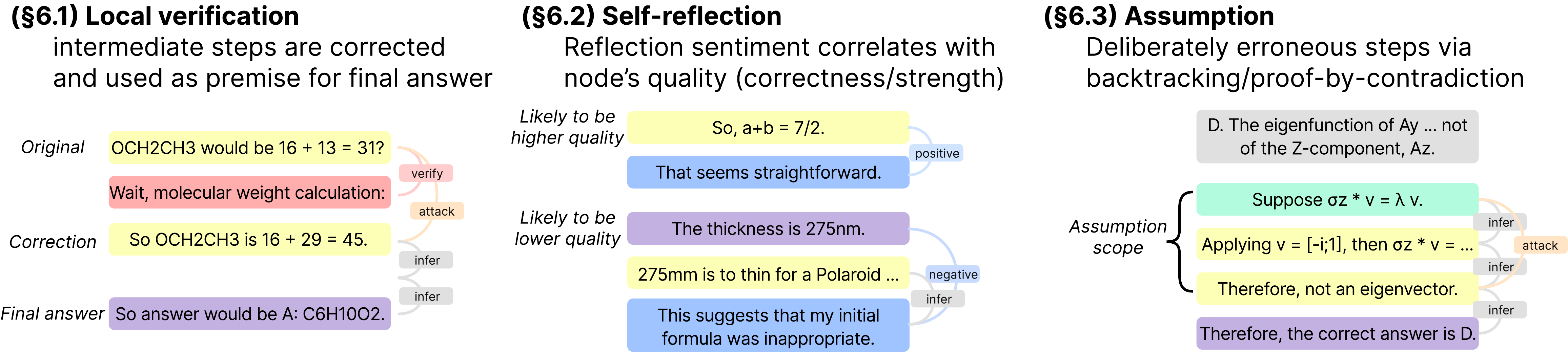}
    \caption{Examples of three fine-grained reasoning behaviors (local verification, self-reflection, and assumption).}
    \label{fig:reasoning-behaviors}
\end{figure*}

\section{ReasoningFlow and reasoning behaviors}
\label{sec:behaviors}


\subsection{Local verification}
\label{subsec:local-verification}

\begin{tcolorbox}[findingsbox, title={\textcolor{white}{Finding 2.1}}]
  Local verification effectively corrects erroneous steps before the final answer is derived.
\end{tcolorbox}

\begin{figure}
    \centering
    \includegraphics[width=\linewidth]{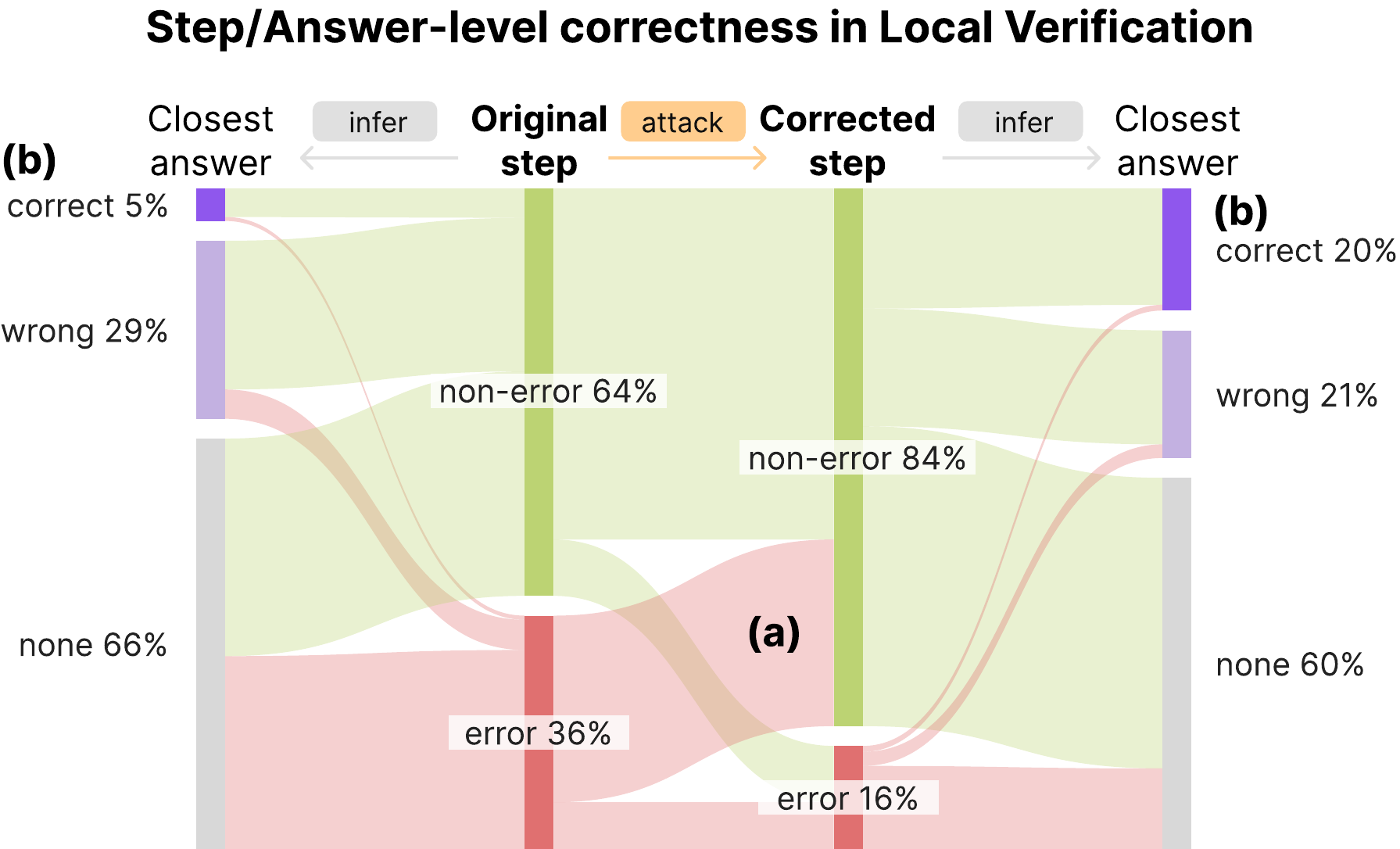}
    \caption{Sankey plot showing how local verification with explicit correction (\edgevalidateattack) affects subsequent reasoning. (a) Local verification, when used, corrects step-level errors (64\% $\rightarrow$ 84\%). (b) Corrected steps are more likely to be used to derive the correct final answer than original steps (20\% vs. 5\%).}
    \label{fig:local-global-verification}
\end{figure}

LRMs engage in self-verification by assessing the correctness of prior reasoning steps and revising them as needed. Previous work focused on \textit{global} verification, where the LRM starts verification on the entire solution after finding the first answer \citep{gandhi_cognitive_2025, marjanovic_deepseek-r1_2026}. Most of these global verifications conclude by simply restating the original answer, regardless of its correctness \citep{zhao_can_2025, liao_lost_2025}.

However, these studies overlook \textbf{local verification}, in which the model detects a potential error mid-reasoning and corrects it within the next few steps (Figure \ref{fig:reasoning-behaviors}). Local verification is universal across different LRMs and datasets, and happens more frequently than global verification of the final answer (19.1 times/trace vs. 6.1 times/trace).

We further analyze local verification cases where the model explicitly \textit{corrects} previous nodes (\edgevalidateattack~edge). These corrections are more likely to be correct reasoning steps (Figure \ref{fig:local-global-verification}(a)), and are significantly more likely to be used to derive the correct final answer(\nodeconclusion~node) than their original statements (Figure \ref{fig:local-global-verification}(b)). This observation shows that local verifications directly steer the following reasoning process towards the correct answer. Improving local verification in LRMs could improve both efficiency, by correcting errors before a first final answer is reached, and monitorability, by localizing where those corrections occur.

\subsection{Self-reflection}
\label{subsec:self-reflection}

\begin{tcolorbox}[findingsbox, title={\textcolor{white}{Finding 2.2}}]
  Self-reflection sentiment correlates with the node quality, enabling a new direction for reasoning monitorability.
\end{tcolorbox}

\begin{figure}
    \centering
    \includegraphics[width=\linewidth]{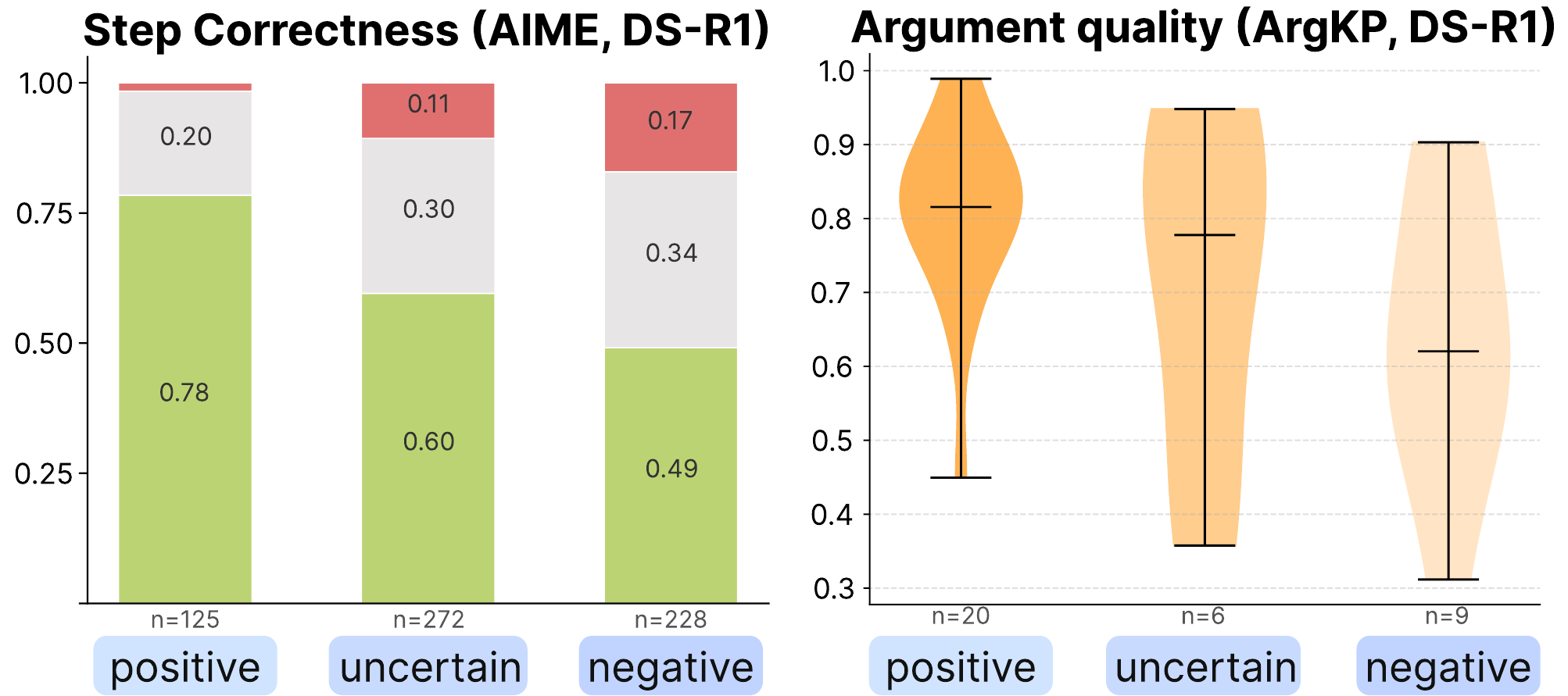}
    \caption{Self-reflection sentiment correlates with node quality (correctness, argument quality). For step correctness, green/gray/red corresponds to correct nodes, propagated errors, and direct errors, respectively (Appendix \ref{sec:appendix-node-quality}).}
    \label{fig:self-reflection}
\end{figure}

\begin{figure*}[ht]
    \centering
    \includegraphics[width=\linewidth]{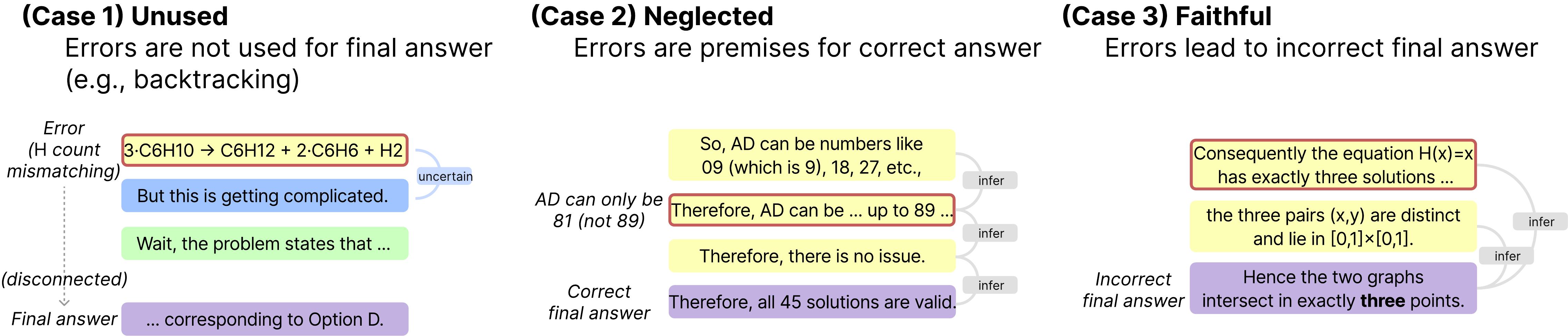}
    \caption{Examples of three types of error handling in LRMs (unused, neglected, and faithful error propagation).}
    \label{fig:validity-eval-examples}
\end{figure*}

Self-reflection consists of meta-evaluation of previous nodes (\textit{"Looks good"}) or emotions/impressions (\textit{"but I am not sure."}). Three edge labels in ReasoningFlow (\edgereflectpositive, \edgereflectuncertain, \edgereflectnegative) capture this behavior along with the reflection sentiment. With ReasoningFlow, we analyze whether LRMs' self-reflection faithfully reflects the quality of the reflected statement.

We observe a clear trend between node quality and self-reflection sentiment in reasoning models. Across three LRMs in AIME/GPQA, nodes with \edgereflectpositive~reflection are 78.1\% correct, while step correctness decreases to 66.2\% in \edgereflectuncertain~and to 45.6\% in \edgereflectnegative.
Similar trends are observed in argumentation quality; positive reflections are associated with stronger arguments, while uncertain/negative reflections indicate weaker arguments.
Overall, this indicates that self-reflection phrases are not just fillers \citep{wang_wait_2025} but reflect the node quality, opening up a new possibility for monitoring internal beliefs of LRMs.

\subsection{Assumption}
\label{subsec:assumption}

\begin{tcolorbox}[findingsbox, title={\textcolor{white}{Finding 2.3}}]
  ReasoningFlow identifies proof-by-contradiction and DFS, enabling evaluation of hypothetical reasoning abilities.
\end{tcolorbox}

Assuming facts that might not be true is a hallmark of human intelligence \citep{van_hoeck_cognitive_2015}. In reasoning, the most popular usage of assumptions is \textit{proof-by-contradiction}: deriving a contradiction from a deliberately false assumption. Another common use case is \textit{depth-first search (DFS)}: when encountering mutually exclusive branches, assuming one option at a time instead of considering all options simultaneously.

In ReasoningFlow, proof-by-contradiction is realized by \nodeassumption--\edgevalidateattack$\rightarrow$\nodereasoning\ triplets that show assumption being refuted by following nodes (Figure~\ref{fig:reasoning-behaviors}), while DFS is most frequently characterized by \nodeassumption--\edgeplanproceed$\rightarrow$\nodeassumption~triplets that chain multiple assumptions together to enable exhaustive search. ReasoningFlow identifies 161 proof-by-contradiction and 2,514 DFS-related patterns, covering 41.9\% of the total \nodeassumption\ nodes.

Deliberate errors that follow a false assumption should be treated differently from normal errors. Evaluating the hypothetical reasoning ability of LLMs, not by the truth value of the statement but by the plausibility of the assumption or the exhaustiveness of DFS, remains underexplored for evaluating higher-order reasoning of LRM.

\section{ReasoningFlow and stepwise evaluation}
\label{sec:validity-evaluation}

\begin{tcolorbox}[findingsbox, title={\textcolor{white}{Finding 3}}]
  Most erroneous steps in LRMs do not lead to incorrect final answers.
\end{tcolorbox}

Stepwise evaluation identifies erroneous steps within a reasoning trace, often using classifiers (Process Reward Models) or LLM-as-a-judge \citep{lee_evaluating_2025}. Early works implicitly assumed that a logical error leads to an incorrect final answer, e.g., Best-of-N sampling that filters out any traces containing errors \citep{ling_deductive_2023, lightman_lets_2024}. However, empirical evidence suggests that LLMs can produce erroneous intermediate steps and yet still arrive at correct conclusions \citep{yee_dissociation_2024, kim_scaling_2025}, likely because not all steps are used to derive the final answer.

Since ReasoningFlow tracks the premises of every step, we can determine whether a given step contributed to the final answer. Figure \ref{fig:validity-eval-examples} illustrates three possible relations between an error and the final answer: the error plays no role in any subsequent conclusion (\textit{Unused}), the error is used as a premise of a correct answer (\textit{Neglected}), or the error propagates to produce an incorrect answer (\textit{Faithful}).

ReasoningFlow reveals that in LRMs, only 14.4\% of the erroneous nodes propagate to incorrect final answers. First, 79.6\% of the errors do not connect to the final answer, indicating that most errors are \textit{unused} during reasoning. This occurs particularly during backtracking, when the LLM did not develop a final answer while exploring that direction. Second, the remaining 6.0\% of the errors (32.8\% of the errors connected to \nodeconclusion~nodes) lead to a \textit{correct} final answer, indicating that the error was \textit{neglected} within the core arguments. Detailed statistics for each (model, dataset) pair are presented in Appendix \ref{sec:appendix-validity-evaluation}.

These unused and neglected errors together prove that the logical link between reasoning errors and incorrect final answers is often weak, highlighting the unfaithfulness of LRMs in handling errors. We argue that stepwise evaluation can be improved by incorporating the discourse structure of traces to track how errors propagate downstream.

\section{ReasoningFlow and mechanistic interpretability}
\label{sec:thought-anchors}

\begin{tcolorbox}[findingsbox, title={\textcolor{white}{Finding 4}}]
  Mechanistically measured causal step-to-step dependencies do not align with language-level discourse relations.
\end{tcolorbox}

\paragraph{Mechanistic interpretation of trace structures.} Thought Anchors \citep{bogdan_thought_2025} proposes measuring the causal dependency between reasoning steps by masking attentions. To measure the dependency between steps $s_i$ and $s_j$ ($i < j$), it first obtains $s_{j}$'s token probability distribution. Then, it recomputes the token probabilities after masking all tokens in $s_i$. Finally, the causal dependency score is defined as the average log KL-divergence between the probability distribution over all tokens in step $s_j$. Qualitatively, a higher score implies that $s_i$ casts a significant effect on the generation of $s_j$, while a lower score implies the opposite.

\begin{figure}
    \centering
    \includegraphics[width=\linewidth]{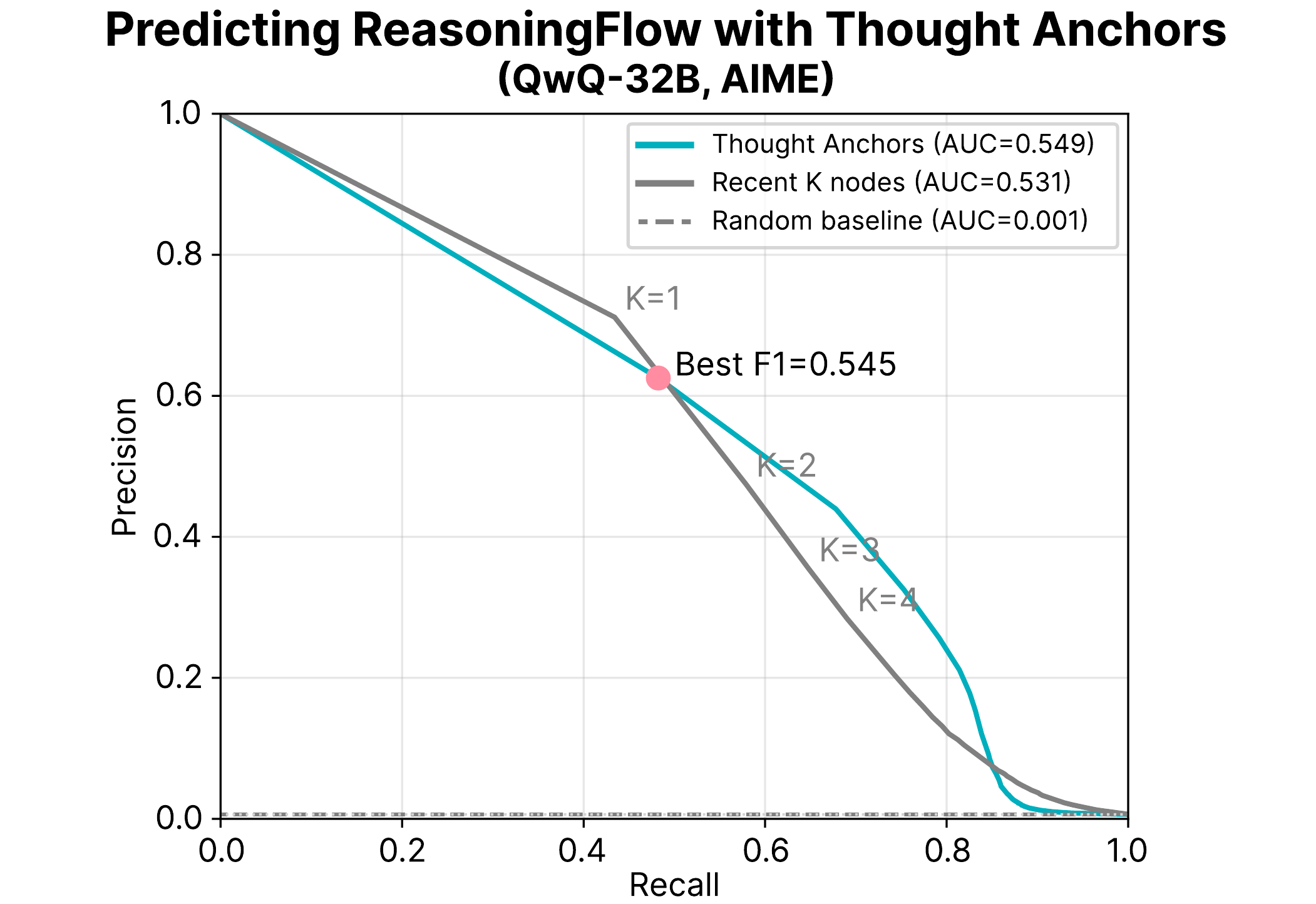}
    \caption{Precision-Recall curve of using Thought Anchors' scores to predict ReasoningFlow edges. Full results are presented in Figure \ref{fig:thought-anchors-full}.}
    \label{fig:thought-anchors-small}
\end{figure}

\paragraph{Results.}
Figure \ref{fig:thought-anchors-small} shows the precision-recall graph of using Thought Anchors' scores to predict ReasoningFlow edges in (QwQ, AIME). While Thought Anchors outperform the random baseline, they are not significantly better compared to greedily choosing the closest $K=1,2,3,...$ nodes. This indicates that the distance between two nodes is the primary driver of the positive correlation.

Comparing Thought Anchors and ReasoningFlow reveals the discrepancy between the text-level structures and their mechanistic interpretation. This aligns with prior work on synthetic logical benchmarks \citep{zhong_chains_2026} and in-context retrieval \citep{du_context_2025}, which found that greater distance between the steps alone introduces significant noise into semantic dependency representations. Aligning internal representations with surface semantics is a crucial challenge for faithful LRMs, where ReasoningFlow can serve as a ground truth for human perception of the language reasoning trace.

\section{Conclusion}

We propose ReasoningFlow, a comprehensive framework for annotating discourse structures in reasoning traces. ReasoningFlow dataset comprises 1.3k manually and automatically annotated traces with fine-grained node and edge labels. ReasoningFlow can be used to show structural similarity between LRMs, discover novel reasoning behaviors to the sub-sentence level, assess how erroneous steps causally affect the final answer, and identify the gap between mechanistic and discourse structures in reasoning models.

Beyond the analyses presented here, we believe ReasoningFlow can serve as a general, human-interpretable lens for studying the reasoning capabilities of LRMs. Promising directions include examining how reasoning behaviors emerge over the course of training \citep{ettinger_olmo_2025, wang_emergent_2025}, how LRMs adapt their reasoning under constrained inference-time compute budgets \citep{wen_budgetthinker_2025}, and how they resolve conflicting knowledge \citep{lin_resisting_2025}.

\section*{Limitations}

ReasoningFlow does not include manual verification of \textit{all} data (1.2k traces) due to significant annotation cost. To ensure the quality of the data and downstream analyses, we present partial manual verification results throughout the data annotation pipeline (Section \ref{sec:dataset-construction}) in Appendix \ref{sec:appendix-automatic-annotation}-\ref{sec:appendix-node-quality}.

ReasoningFlow includes three representative open-weight LRMs (QwQ, DeepSeek-R1, GPT-oss). Closed-source models (e.g., o1, Gemini, Claude) or distilled models (e.g., DeepSeek-R1-Distill \citep{guo_deepseek-r1_2025}, OpenThoughts \citep{guha_openthoughts_2025}) might demonstrate different behaviors when analyzed with ReasoningFlow. However, the fact that independently trained reasoning models exhibit structural similarity (Section \ref{sec:statistics}) suggests that other LRMs will likely exhibit a similar distribution.

The current automatic annotation pipeline for ReasoningFlow is costly. At this moment, we prioritize the dataset quality over the inference cost by using API-based frontier models. The automatically annotated data can be used to train smaller, specialized models, which will significantly reduce the annotation cost and increase the availability of ReasoningFlow.

Furthermore, we emphasize that many downstream tasks would not need full ReasoningFlow annotations. For instance, to distinguish between reasoning and non-reasoning parts (e.g., \nodereflection), one can only use the node annotations; to identify 

\section*{Acknowledgements}

We gratefully acknowledge LBOX for their generous research gift. We also thank Sagnik Mukherjee, Yukyung Lee, and Takyoung Kim for their valuable discussions and insights.

\bibliography{custom}

\appendix
\onecolumn
\section{ReasoningFlow annotation guide}
\label{sec:appendix-annotation-guide}

This section includes annotation guides for nodes and edges. For each subtype of nodes and edges, we provide one example to facilitate understanding of the presented labels due to spatial limitations. In the actual annotation guides for the manual annotators (Section~\ref{sec:manual-annotation}) and LLM prompts, we provide the full set of examples, including an average of 2.05 examples per subtype.

\small
\begin{longtable}{@{} L{2.4cm} L{5cm} L{2.5cm} L{4.5cm} @{}}
\toprule
\textbf{Label} & \textbf{Description} & \textbf{Subtype} & \textbf{Example} \\
\midrule
\endfirsthead
\toprule
\textbf{Label} & \textbf{Description} & \textbf{Subtype} & \textbf{Example} \\
\midrule
\endhead
\midrule
\multicolumn{4}{r}{\textit{Continued on next page}} \\
\endfoot
\bottomrule
\endlastfoot
\colorbox{nodecontext}{\strut\texttt{Context}} & \textbf{Context} includes all \textit{user-provided} texts. Some examples include the problem statement, retrieved documents, and tool responses. & &  \\[4pt]
\midrule
\colorbox{nodeplanning}{\strut\texttt{Planning}} & \multirow{5}{5cm}{\textbf{Planning} introduces the content of the following nodes. It can be both coarse, high-level direction that affects tens to hundreds of nodes, or highly local plans that only affect the next node. Planning node often includes phrases like (I need, I should, Let's, ...) indicating that the model should \textbf{act} in some way. However, it can also be a very short phrase that indicates the direction of the next node, such as \textit{"Now, for n $\geq$ 2:"} or \textit{"Numerator:"}.} & Introducing long-term directions and subgoals (global plans) & First, I need to remember what Gibbs free energy change, $\Delta$G, represents. \\
 \cmidrule{3-4} &  & Phrases that initiate verification & I should also consider if there are any assumptions I'm making here. \\
 \cmidrule{3-4} &  & Phrases that initiate backtracking (alternative solutions) & Alternatively, maybe I can think in terms of permutations with restrictions. \\
 \cmidrule{3-4} &  & Introducing the direction of the next node (local plans) & Let me calculate the numerator: \\
 \cmidrule{3-4} &  & Indicating the Conclusion node & Final Answer: \\
\midrule
\colorbox{nodefact}{\strut\texttt{Fact}} & \multirow{5}{5cm}{\textbf{Fact} contains external, parametric knowledge that is independent from the information provided in the context. A fact node must satisfy two criteria: 1. The information should not be given in any of the previous nodes including contexts. If so, it should be considered as restatement nodes. 2. It must not include nor reference information (numeric values, propositional content, context-specific constraints, ...) from the previous steps. If so, it should be considered as reasoning nodes.} & Theorems, laws, and rules & E° cell = E° cathode + E° oxidation. \\
 \cmidrule{3-4} &  & Existing concepts and definitions & The Legendre symbol \(\left( \frac{a}{p} \right)\) is defined as 1 if \(a\) is a quadratic residue modulo \(p\) and not divisible by \(p\), and -1 otherwise. \\
 \cmidrule{3-4} &  & Values of well-known constants and unit conversions & I think A for nitrogen at 77 K is a known value. \\
 \cmidrule{3-4} &  & Factual knowledge & Now, hydrochloric acid (HCl) is a strong acid, which means it completely dissociates in water. \\
 \cmidrule{3-4} &  & Commonsense facts & 2 \textasciicircum 10 is 1024. \\
\midrule
\colorbox{nodereasoning}{\strut\texttt{Reasoning}} & \multirow{6}{5cm}{\textbf{Reasoning} includes any of deductive/inductive/abductive inference. Reasoning nodes should not present novel facts; they must refer to previous nodes to derive the designated conclusion. Even if the derivation is trivial or obvious, it should still be categorized as Reasoning as long as it involves logical steps based on prior information.} & Calculations & So, $\Delta$G° = (-2150.4) - (-103.8) = -2150.4 + 103.8 = -2046.6 kJ/mol \\
 \cmidrule{3-4} &  & Logical reasoning & So, there are no real solutions to the inequality. \\
 \cmidrule{3-4} &  & Commonsense reasoning & When we increase the tax imposed on cigarettes, it will reduce the demand. \\
 \cmidrule{3-4} &  & Speculations & This logarithmic relationship suggests that the first digit changes in a way that is not periodic. \\
 \cmidrule{3-4} &  & Defining symbols and notations & Let the speed of the spacecraft be \(v\). \\
 \cmidrule{3-4} &  & Observations from examples & It seems that the provided numbers are all prime numbers. \\
\midrule
\colorbox{noderestatement}{\strut\texttt{Restatement}} & \multirow{1}{5cm}{\textbf{Restatement} is when the model copies/paraphrases the preceding text (context/previous nodes). Therefore, the destination node should have content that can be directly entailed from the source node. If the node includes any new information that is not present in the source node, it should not be classified as Restatement. Typically, terms like 'as seen previously', 'already stated', 'the problem', strongly indicate the restatement nodes. However, it might simply restate previous steps without any of these auxiliary phrases.} & Rephrasing the context nodes & The temperature is 25°C and the pressure is 1 atm. \\[2cm]
 \cmidrule{3-4} &  & Rephrasing previous nodes & As before, this is approximately 13.928, which rounds to 14. \\[2cm]
\midrule
\colorbox{nodeassumption}{\strut\texttt{Assumption}} & \multirow{5}{5cm}{\textbf{Assumption} is when a step is intentionally indicated as \textit{not necessarily true}, but serves as a premise for the following steps. Therefore, Assumption defines a \textit{scope} of nodes that are based on the assumption, where the validity of nodes depends on the validity of the assumption.} & Assuming missing premises by commonsense/common practices & Since it's a swimming pool, I assume it's open, \\
 \cmidrule{3-4} &  & Assuming uncertain facts & Assuming the C-H stretching wavenumber in methane is around 3000 cm$^{-1}$, \\
 \cmidrule{3-4} &  & Branching (switch-case) & If it were 4, ... Otherwise, if it were 6, ... \\
 \cmidrule{3-4} &  & Proof by contradiction & Let's suppose, for the sake of contradiction, that there exists an integer \(m\) such that \(m^2 = 4n + 3\) for some integer \(n\). \\
\midrule
\colorbox{nodeexample}{\strut\texttt{Example}} & \multirow{3}{5cm}{\textbf{Example} is a specific instance of a general and abstract concept provided in the preceding steps. Examples are often unnecessary for the reasoning process, but rather helps illustrate the outline of the solution. Therefore, even if it is enumerating multiple objects, if it is directly solving the problem, it should be categorized as Reasoning. If a single example (each of the examples in a list) spans multiple nodes, label the first one as example, and the following nodes (derived from the first node) as accordingly.} & Pattern induction by enumeration & Take \( n = 1 \): \\[1.1cm]
 \cmidrule{3-4} &  & Non-exhaustive listings & Suppose \(s = 13\), \(s^2 = 169\), sum is \(\frac{169 - 95}{2} = \frac{74}{2} = 37\), which is larger than \(13\). \\[1.1cm]
 \cmidrule{3-4} &  & Rhetorical examples & For example, governments can provide services for verifying one's identity for financial transactions, preventing misuse. \bigstrut \\[1.1cm]
\midrule
\colorbox{nodereflection}{\strut\texttt{Reflection}} & \multirow{5}{5cm}{\textbf{Reflection} is a node that expresses \textit{opinions} on the previous nodes. Reflections include both illogical emotions like \textit{curiosity, uncertainty, and satisfaction}, and logical judgments like \textit{correctness}. However, LLMs sometimes include subjective words during the reasoning process, as in "I think that (fact)" or "Wait, (fact)". Make sure the reflection nodes are not directly a part of the solution, i.e., removing the reflection node will make the reasoning trace still logically complete.} & Meta-evaluation of a step & I'm clearly making a mistake here. \\
 \cmidrule{3-4} &  & Emotions and impressions & This is confusing. \\
 \cmidrule{3-4} &  & Rhetorical phrases & I'm not sure what it is off the top of my head. \\
 \cmidrule{3-4} &  & Filler words & I got this problem here. \\
 \cmidrule{3-4} &  & Reasoning on the applicability of Fact & However, Benford's law itself doesn't directly help in proving non-periodicity. \\
\midrule
\colorbox{nodeconclusion}{\strut\texttt{Conclusion}} & \multirow{3}{5cm}{This node includes the model's answer for the question (both final/intermediate). Final answer-based conclusion nodes are the ones that appear on the last attempt, which often includes phrases like \textbf{Final Answer}, \textbf{Answer:}, or similar expressions. However, conclusion nodes can contain \textit{intermediate} answers. These conclusion nodes are usually the last step of an "attempt" to solve the problem, and are often followed by planning nodes that init verification. Exclude nodes that are scoped under assumptions. Make sure that the model is asserting the conclusion node as the potential final answer.} & Intermediate conclusions & \- \textit{(Question is asking for the time dilation factor)} Total time dilation factor: 0.9851 / 1.6667 $\approx$ 0.591. \\[2.5cm]
 \cmidrule{3-4} &  & Final answers & Final answer is 5. \\[2.5cm]
\caption{Annotation guide for node labels.}
\label{tab:node-labels}
\end{longtable}
\normalsize

\small
\begin{longtable}{@{} L{2.4cm} L{5cm} L{2.5cm} L{4.5cm} @{}}
\toprule
\textbf{Edge} & \textbf{Description} & \textbf{Subtype} & \textbf{Example} \\
\midrule
\endfirsthead
\toprule
\textbf{Edge} & \textbf{Description} & \textbf{Subtype} & \textbf{Example} \\
\midrule
\endhead
\midrule
\multicolumn{4}{r}{\textit{Continued on next page}} \\
\endfoot
\bottomrule
\endlastfoot
\colorbox{edgereasoninfer}{\strut\texttt{infer}} & \multirow{5}{4cm}{This edge represents the basic premise-conclusion relationship. To apply this edge, both nodes should be asserted facts/propositions. There might be multiple premise nodes that are semantically equivalent. (1) If it is due to restatement, choose only the original node instead of the restatement. (2) Otherwise, choose the closest source node from the destination.} & (Deductive) Premises → Conclusion & \texttt{fact}: \textit{Sulferic acid is a strong acid,} $\Rightarrow$ \texttt{reasoning}: \textit{so it will fully dissociate, providing H+ ions.} \\
 \cmidrule{3-4} &  & (Math) Inputs and equations → Calculation results & \texttt{reasoning}: \textit{r \textasciicircum 2 h = 2r \textasciicircum 3} $\Rightarrow$ \texttt{reasoning}: \textit{h = 2r} \\
 \cmidrule{3-4} &  & (Inductive) Observations → Inductive Hypothesis & \texttt{example}: \textit{7 \textasciicircum 2=49, which is 4 * 12 + 1.} $\Rightarrow$ \texttt{reasoning}: \textit{Hmm, it seems like all these perfect squares are either 4n+0 or 4n+1.} \\
 \cmidrule{3-4} &  & Rationales for self-reflection & \texttt{reasoning}: \textit{Wait, water cannot be liquid at 1000 K, 1 atm.} $\Rightarrow$ \texttt{reflection}: \textit{This seems off.} \\
\midrule
\colorbox{edgereasonexecute}{\strut\texttt{execute}} & This label represents the relationship between a plan and its execution. The plan "introduces" what will happen in the subsequent reasoning step, and the subsequent nodes will follow the plan to perform actual reasoning. Note that this edge is mostly short-distanced, where it often connects nodes that are within distance 2. If there are parallel reasoning going on (e.g., by individually assessing two possibilities), this edge should connect the plan and the first nodes of these parallel reasoning chains. A step might be logically correct, but might not be relevant to the current goal. This edge provides the contextual information regarding what the step is doing right now, i.e., coherence of the reasoning step. & A plan and its implementation & \texttt{planning}: \textit{Let me consider some examples:} $\Rightarrow$ \texttt{example}: \textit{- 0 \textasciicircum 2 = 0, which is 4×0 + 1.} \\
\midrule
\colorbox{edgereasonrestate}{\strut\texttt{restate}} & \multirow{2}{5cm}{This edge connects the Restatement node to the previous node that is restated. When there are multiple statements that include the same content, make sure that this edge points to the original node that is the earliest in the reasoning trace. Even if the node is not directly copying the content, restating a few facts or high-level information of the context can be considered as restatement.} & Restating contexts & \texttt{context}: \textit{What is the amplitude of gravitational waves produced by a binary black hole system with masses of 10 solar masses and 25 solar masses at a distance of 500 megaparsecs from the Earth?} $\Rightarrow$ \texttt{restatement}: \textit{Alright, I'm trying to figure out the amplitude of gravitational waves from a binary black hole system.} \\
 \cmidrule{3-4} &  & Restating intermediate nodes & \texttt{assumption}: \textit{Let’s suppose, for the sake of contradiction, that there exists an integer m such that $m^2 = 4n + 3$ for some integer $n$.} $\Rightarrow$ \texttt{restatement}: \textit{I previously assumed that an integer $m$ satisfies $m^2 = 4n + 3$.} \\
\midrule
\colorbox{edgereasonelaboratefact}{\strut\texttt{elaborate-fact}} & This edge connects general facts to more specific facts. This edge is often used when the LLM generates consequent nodes of facts, and the following nodes describe the details of the initial fact. Note that this edge is usually short-distanced, where it is often connecting nodes that are within distance 2. Even if the two facts are semantically related, if they are not introduced within the same context, they should be not connected.' & A fact and linked facts & \texttt{fact}: \textit{Okay, so I remember that the Arrhenius equation relates these quantities.} $\Rightarrow$ \texttt{fact}: \textit{The equation is k = A * e \textasciicircum (-Ea/(RT)),} \\
\midrule
\colorbox{edgereasonexemplify}{\strut\texttt{exemplify}} & \multirow{3}{5cm}{This edge label applies to Example nodes, connecting the example to what it exemplifies. When both nodes are trivia-like facts, it should be rather annotated as fact-detail.} & Enumeration & \texttt{planning}: \textit{First, maybe I should think about what perfect squares look like when divided by 4.} $\Rightarrow$ \texttt{example}: \textit{- 0 \textasciicircum 2 = 0, which is 4×0 + 0.} \\
 \cmidrule{3-4} &  & Conceptual examples & \texttt{planning}: \textit{Maybe the problem is considering some other reaction or there's additional information I need to account for.} $\Rightarrow$ \texttt{example}: \textit{Perhaps the presense of sulfuric acid allows for copper to be oxidized directly, or there's a coupled reaction happening that I'm not accounting for.} \\
 \cmidrule{3-4} &  & Demonstrative examples & \texttt{reasoning}: \textit{Therefore, the valid range is 0 < c < 9.} $\Rightarrow$ \texttt{example}: \textit{Take c=4, which is between 0 and 9.} \\
\midrule\pagebreak
\colorbox{edgeplanproceed}{\strut\texttt{proceed}} & \multirow{6}{5cm}{This is the default label that connects a plan to any nodes which the content triggers the planning. Typical examples include: - deciding high-level plans based on known facts - indicating how to solve (simplify, substitute, calculate) a certain equation at a low level - multiple planning/assumption nodes forming a linear sequence (first, second, .../ next, then, ...) verify, decompose, and backtrack edges are exceptions; they should be annotated with higher priority than this label. A rule of thumb for determining source nodes that are not planning: if a plan leads to a reasoning node via "execute" edge, the relevant source nodes of that reasoning node (annotated as infer/execute) should be connected to the plan via this edge. A rule of thumb for determining planning source nodes: the planning must connect to the very next planning node in a linear chain.} & Steps that motivate the next plan/assumption → a plan/assumption & \texttt{reasoning}: \textit{I don't see any patterns in the sequence.} $\Rightarrow$ \texttt{assumption}: \textit{Let's assume that the sequence is not periodic.} \\
 \cmidrule{3-4} &  & The previous plan's conclusion → next plan & \texttt{reasoning}: \textit{So, $M_c$ = 250 \textasciicircum (3/5) / 35 \textasciicircum (1/5).} $\Rightarrow$ \texttt{planning}: \textit{I need to calculate that.} \\
 \cmidrule{3-4} &  & Linear progression of planning & \texttt{planning}: \textit{Left side:} $\Rightarrow$ \texttt{planning}: \textit{Right side:} \\
 \cmidrule{3-4} &  & Mutually exclusive assumptions & \texttt{assumption}: \textit{Let's consider an even number: let's say 2k, where k is an integer.} $\Rightarrow$ \texttt{assumption}: \textit{Now, consider an odd number: let's say 2k+1, where k is an integer.} \\
 \cmidrule{3-4} &  & The reason a plan is not working → alternative plan & \texttt{reasoning}: \textit{But I don't have a specific frequency given in the problem.} $\Rightarrow$ \texttt{planning}: \textit{Maybe I need to use the frequency at a certain point in the waveform, like the frequency at merger.} \\
 \cmidrule{3-4} &  & Steps that motivate self verification → Planning node that initializes verification & \texttt{reflection}: \textit{This seems off.} $\Rightarrow$ \texttt{planning}: \textit{Maybe there's something missing.} \\
\midrule
\colorbox{edgeplanverify}{\strut\texttt{verify}}& This edge specifically labels the relationship between a reasoning-like node and a Planning node, initiating the verification process. Typical destination nodes include phrases like "Let's verify this", "I should check if there is any error". & Initiating verification & \texttt{reasoning}: \textit{So, a + b = 7/2.} $\Rightarrow$ \texttt{planning}: \textit{But let me double-check.} \\
\midrule
\colorbox{edgeplandecompose}{\strut\texttt{decompose}} & \multirow{3}{5cm}{This edge connects a plan and a subplan. Subplans decompose the high-level plan into smaller, more straightforward goals. If the plans can be executed in parallel, edges should connect the high-level plan to all subplans; if the subplans are meant to execute sequentially, edges should connect to the first plan only. This is the most standard relationship between plans, so if there is a clear semantic relationship between two plans but no strong evidence for proceed or backtrack, this edge should be used.} & Decomposition of coarser plans into finer subplans & \texttt{planning}: \textit{Let's compute this equation step by step.} $\Rightarrow$ \texttt{planning}: \textit{First, computing 93.51 × 123.7:} \\
 \cmidrule{3-4} &  & Dividing a coarse assumption into fine-grained ones & \texttt{assumption}: \textit{Let's assume that n \textasciicircum 2 is either 0 or 1 modulo 4.} $\Rightarrow$ \texttt{assumption}: \textit{Case 1: n \textasciicircum 2 $\equiv$ 0 (mod 4)} \\
 \cmidrule{3-4} &  & Paraphrasing of a planning node due to formatting reasons (should only select when two nodes are adjacent) & \texttt{planning}: \textit{Let's consider the case for x=5.} $\Rightarrow$ \texttt{planning}: \textit{\#\#\# Case 1: x=5} \\
\midrule
\colorbox{edgeplanbacktrack}{\strut\texttt{backtrack}} & This edge denotes when the model is suggesting an alternative plan from the previous one. The subsequent nodes often include phrases that indicate backtracking, e.g., "Alternatively", "Let me think in other direction". When there are multiple alternative Planning steps chained linearly, one should only link the immediately preceding one and not connect between all possible pairs. & Alternative plans (backtracking) & \texttt{planning}: \textit{Also, I should consider if there are any other factors that might affect the pressure, like temperature or the exact value of gravity,} $\Rightarrow$ \texttt{planning}: \textit{Another thing to think about is whether the pool is open to the atmosphere or not.} \\
\midrule
\colorbox{edgereflectpositive}{\strut\texttt{positive}} & Reflection edges connect a planning/reasoning-like statement to a reflection node that contains personal views and emotions about the statement. This edge connects a reasoning node (mostly Reasoning, Fact, Assumption) and a reflection node that affirms it. If the affirmation is based on the semantic identity of previous and current steps, e.g., connecting the conclusion of the positive self-verification to the verification goal, it should be annotated as support. If the reflection node compares/contrasts multiple nodes (e.g., "This seems consistent with previous results"), create edges for both nodes that are being compared. & Affirmation of previous nodes & \texttt{conclusion}: \textit{H2SO4 + Ca(OH)2 → CaSO4 + 2 H2O} $\Rightarrow$ \texttt{reflection}: \textit{I think that's it.} \\
\midrule
\colorbox{edgereflectnegative}{\strut\texttt{negative}}& \multirow{3}{5cm}{Reflection edges connect a planning/reasoning-like statement to a reflection node that contains personal views and emotions about the statement. This edge connects a reasoning node and a node that denies it. This is mostly caused by reflection statements like "This seems incorrect" or reflecting without explicit correction edge, such as "Therefore, it is wrong". When it explicitly corrects the previous node, it should be annotated as attack. If the reflection node compares/contrasts multiple nodes (e.g., "This seems off given the previous results"), create edges for both nodes that are being compared.} & Negative evaluation of previous reasoning results & \texttt{reasoning}: \textit{- H: 2 + 2*2 = 6?} $\Rightarrow$ \texttt{reflection}: \textit{Wait, no.} \\[1.5cm]
 \cmidrule{3-4} &  & Negative judgment on the applicability of a fact & \texttt{fact}: \textit{Van der waals forces are weak intermolecular forces that arise from induced dipoles in molecules.} $\Rightarrow$ \texttt{reflection}: \textit{However, I think that it will be negligible in this case.} \\[1.5cm]
 \cmidrule{3-4} &  & Negative judgment on whether the plan will work or not & \texttt{planning}: \textit{We can try considering Van der waals forces.} $\Rightarrow$ \texttt{reflection}: \textit{However, I think that it will be negligible in this case.} \\[1.5cm]
\midrule
\colorbox{edgereflectuncertain}{\strut\texttt{uncertain}} & \multirow{5}{5cm}{Reflection edges connect a planning/reasoning-like statement (source) to a reflection node (dest) that contains personal views and emotions about the statement. This edge represents all reflections that are not explicitly positive or negative. Typical use cases are reflection of uncertainty ("I am not sure/certain"), confusion ("I am confused/lost"), lack of knowledge ("I don't know"), lack of confidence ("I am a bit rusty"), and anomaly ("This seems weird"). If the reflection node compares/contrasts multiple nodes (e.g., "This is different from the previous result, which confuses me"), create edges for both nodes that are being compared.} & Uncertainty & \texttt{planning}: \textit{But wait, Is this formula correct?} $\Rightarrow$ \texttt{reflection}: \textit{I'm not entirely sure.} \\
 \cmidrule{3-4} &  & Confusion & \texttt{reasoning}: \textit{t = 550 nm / 1.5 = 366.67 nm.} $\Rightarrow$ \texttt{reflection}: \textit{This is different from the previous results.} \\
 \cmidrule{3-4} &  & Lack of knowledge/ability & \texttt{reasoning}: \textit{t = 550 nm / 1.5 = 366.67 nm.} $\Rightarrow$ \texttt{reflection}: \textit{But again,. without a clear theoretical basis, it's hard to be confident.} \\
 \cmidrule{3-4} &  & Lack of confidence; too difficult & \texttt{reasoning}: \textit{but again, that's for reflections, not necessarily for transmission through a material.} $\Rightarrow$ \texttt{reflection}: \textit{This is tricky.} \\
 \cmidrule{3-4} &  & Exclamation & \texttt{reasoning}: \textit{so total Ea = 94,147.736 + 691.33 $\approx$ 94,839 J/mol.} $\Rightarrow$ \texttt{reflection}: \textit{That seems high.} \\
\midrule\pagebreak
\colorbox{edgevalidatesupport}{\strut\texttt{support}} & \multirow{3}{5cm}{While other edges are focused on the logical premise and the reasoner's intent, Validate edges denote whether two pieces of information are consistent or not. This edge label applies to reasoning nodes and conclusion nodes that reinforce previous nodes by stating the same thing. It distinguishes from restate because it is not blindly copying the previous statement, but comparing two pieces of information that were derived independently. It also differs from positive, because support annotates propositional equivalence while positive only annotates positive sentiment. Finally, `Conclusion` nodes must connect to all previous `conclusion` nodes that address the same question, via support or attack.} & Positive self-verification conclusion & \texttt{reasoning}: \textit{Now, ln(A/k) = ln(5e13 / 2.5e-3) = ln(2e16) $\approx$ 38.279, as I had before.} $\Rightarrow$ \texttt{reasoning}: \textit{So, I use 38.279, which is close enough.} \\[2cm]
 \cmidrule{3-4} &  & Conclusion node and preceding consistent conclusions & \texttt{conclusion}: \textit{H2SO4 + Ca(OH)2 → CaSO4 + 2 H2O} $\Rightarrow$ \texttt{conclusion}: \textit{H2SO4 + Ca(OH)2 → CaSO4 + 2 H2O} \\[2cm]
 \cmidrule{3-4} &  & The context is a proposition, and the conclusion is its affirmation & \texttt{context}: \textit{Should cigarrettes be banned for all ages?} $\Rightarrow$ \texttt{conclusion}: \textit{As my final opinion, I believe that all cigarrettes should be banned for all ages.} \\[2cm]
\midrule
\colorbox{edgevalidateattack}{\strut\texttt{attack}} & \multirow{4}{5cm}{While other edges are focused on the logical premise and the reasoner's intent, Validate edges denote whether two pieces of information are consistent or not. This edge label applies to Restatement nodes and Conclusion nodes, connecting the original statement to the restatement. It differs from negative, because attack annotates propositional equivalence, while negative only annotates positive sentiment. Finally, `Conclusion` nodes must connect to all previous `conclusion` nodes that address the same question, via support or attack.} & Proof by contradiction & \texttt{assumption}: \textit{Let's suppose, for the sake of contradiction, that there exists an integer m suc that m \textasciicircum 2 = 4n + 3 for some integer n.} $\Rightarrow$ \texttt{reasoning}: \textit{it must be that no integer m exists such that m \textasciicircum 2 = 4n + 3.} \\
 \cmidrule{3-4} &  & Explicit correction & \texttt{reasoning}: \textit{H: 2 + 2*2 = 6?} $\Rightarrow$ \texttt{reasoning}: \textit{H2SO4 has 2 H, and Ca(OH)2 has 2*(1) = 2 H, so total 4 H.} \\
 \cmidrule{3-4} &  & A claim and preceding claims that are logically inconsistent/mutually exclusive & \texttt{conclusion}: \textit{The answer is Wednesday.} $\Rightarrow$ \texttt{conclusion}: \textit{The final answer is Tuesday.} \\
 \cmidrule{3-4} &  & The context is a proposition, and the conclusion is its negation & \texttt{context}: \textit{Should cigarrettes be banned for all ages?} $\Rightarrow$ \texttt{conclusion}: \textit{To conclude, I think that cigarrettes should be allowed to adults.} \\
 \caption{Annotation guide for edge labels.}
\label{tab:edge-labels}
\end{longtable}
\normalsize

\twocolumn

\section{Extended related works}
\label{sec:appendix-literature-review}

\subsection{LLM reasoning}

\paragraph{Entailment graph and verification.} Reasoning was traditionally viewed as combining existing pieces of knowledge to deduce new facts. \citet{dalvi_explaining_2021} introduced \textit{entailment graph} structures that connect logical premises to their conclusions. Under this formulation, reasoning becomes tree expansion, where one recursively adds intermediate conclusions from the given leaf nodes (facts) until the desired conclusion is reached. The definition of logical premises is straightforward in some tasks. In arithmetic word problems \citep{cobbe_training_2021}, the premises are where the numbers used in the current step are first derived \citep{li_making_2023}; in logical reasoning tasks with corresponding formal logic representation, \textit{natural deduction} reveals the logical premise to deduce the current step \citep{han_p-folio_2024}.

More recent works \citep{ling_deductive_2023, mukherjee_premise-augmented_2025} developed a general definition of premises based on the \textit{stepwise evaluation} task (judging whether a step is correct or erroneous). In this setting, the premise set should be \textit{complete}, providing sufficient information to determine whether the step is correct or erroneous, and \textit{minimal}, so that removing any premises makes the set incomplete. Premise selection based on such criteria leads to accurate and efficient step verification by removing distractors. ReasoningFlow directly adopts this definition as one of the main design principles, so that the set of connected previous nodes forms a \textit{minimal complete set} as defined above.

However, these works are solely limited to entailment relationships, ignoring diverse discourse relations present in LRM traces beyond entailment. Even before LRMs, behaviors like plan decomposition and backtracking were identified as critical components of LLM reasoning \citep{zhou_least--most_2023, yao_tree_2023}. While these behaviors clearly differ from deductive reasoning, existing works on entailment structure failed to capture these behaviors.

\paragraph{Behavioral analysis of LRMs.} After DeepSeek-R1 gained popularity \citep{guo_deepseek-r1_2025}, immediate attempts to analyze their long and complex reasoning traces involved counting keywords like "Wait" \citep{chang_demystifying_2025} or using few-shot LLMs to classify reasoning behaviors \citep{gandhi_cognitive_2025}. However, these approaches focused on the \textit{existence} of such behaviors, rather than identifying the underlying discourse structure.

DeepSeek-R1 Thoughtology \citep{marjanovic_deepseek-r1_2026} identified the global structure of reasoning traces in four stages. The framework states that LRMs first tend to restate the problem in their own language (\textit{Problem definition}), derive an initial solution (\textit{Bloom}), try recomputation or alternative approaches to verify the initial solution (\textit{Reconstruction}), and decide the final answer (\textit{Final decision}). Their proposed structures are semi-linear; the four stages occur linearly, while the reconstruction stages repeat multiple times. While this observation broadly applies to different reasoning models, it fails to address \textit{nested} behaviors of different granularities (e.g., LLM includes a short self-verification during the Bloom stage). Furthermore, their annotations are based on paragraphs, which disallows fine-grained intent analysis and verification.

LCoT2Tree \citep{jiang_what_2025} and ReJump \citep{zeng_rejump_2025} introduce a tree structure to annotate additional information about where verification and backtracking happen. The Bloom stage in \citet{marjanovic_deepseek-r1_2026} is expressed as a sequence of paragraphs, and any backtracking or verification attempt is shown as branching out from one of the paragraphs. This structure can easily express arbitrary backtracking and verification, which is prominent in tasks involving heavy depth-first search like Game of 24 (e.g., \textit{Use 2,3,5,6 and arithmetic operations to make 24}). While they can express non-entailment relationships and non-linear reasoning structures, they cannot express fine-grained reasoning behaviors, like self-reflection and assumption, which require more expressive node and edge labels.

Thought anchors \citep{bogdan_thought_2025} offer \textit{sentence}-level analyses of reasoning trace structures. They propose eight functional roles of sentences, where some of them directly correspond with ReasoningFlow labels (\textit{Plan generation} and \nodeplanning, \textit{Fact retrieval} and \nodefact, \textit{Deduction/Active computation} and \nodereasoning, \textit{Example testing} and \nodeexample, \textit{Final answer emission} and \nodeconclusion). However, their node annotations focus on \textit{global} roles. For instance, steps within global verification (\textit{Reconstruction} stage) are always annotated as \textit{Self-checking} in Thought Anchors, but vary by their sentence-level semantics in ReasoningFlow. This leads to discrepancies when applying the structures to \textit{downstream tasks}. For example, since we cannot evaluate the validity of non-reasoning nodes like \nodeplanning, we must further distinguish whether a \texttt{Self-checking} sentence contains reasoning content or not.

Edge annotation is also significantly different between Thought Anchors and ReasoningFlow. Thought Anchor's edge annotations were purely mechanistic and did not distinguish between different discourse relations. ReasoningFlow offers a fine-grained taxonomy of the relation between two steps. We also find that annotated edges between the two methods show a significant gap; refer to the main text (Section \ref{sec:thought-anchors}) and Appendix \ref{sec:appendix-thought-anchors}.

Finally, we compare the current version of ReasoningFlow to the \texttt{preview} version \citep{lee_reasoningflow_2025}. The node labels have been updated in several ways: reflective statements that reason about a previous step are now classified as \nodereflection~rather than \nodereasoning, and \nodeconclusion~has been introduced to distinguish the final answer of each attempt from other nodes. For the edge labels, most have been renamed for consistency. We also introduce validate edges (\edgevalidatesupport, \edgevalidateattack) to capture long-range agreements and disagreements between segments, and merge incoming edges of \nodeplanning~from both planning and non-planning nodes into a single label (\edgeplanproceed).

\subsection{Adjacent fields}

While ReasoningFlow's core design is mainly influenced by the LLM reasoning literature, it is also inspired by diverse fields, including computational linguistics, formal logic, and cognitive science.

\subsubsection{Discourse parsing}


\paragraph{Rhetorical Structure Theory (RST)} RST was first proposed by \citet{mann_rhetorical_1988}, aiming to discover the structural organization of long coherent texts. RST constructs a discourse tree of clause spans, where \textit{satellite} spans are recursively adjoined to the \textit{nucleus} span based on the rhetorical intent (\textit{Why did the speaker add this clause?}). This theory led to the development of the RST-DT dataset \citep{carlson_building_2001}, annotating 17 different semantic relations between adjacent spans.

While ReasoningFlow and RST-DT both annotate discourse structures for fine-grained units, they exhibit significant structural differences. RST-DT annotates projective tree structures without crossing edges, while ReasoningFlow uses a directed acyclic graph structure that allows arbitrarily crossing edges. We find that projectivity constraints cannot coexist with the core design principle of ReasoningFlow (edges should capture minimal but \textit{sufficient} context of a step), as approximately 42.6\% of edges should be removed to ensure projectivity in the ReasoningFlow dataset.

In the label set, the most noticeable difference is the granularity of \textit{inference} behaviors. \edgereasoninfer~in ReasoningFlow can correspond to many types in RST-DT: \texttt{background}, \texttt{cause}, \texttt{explanation}, \texttt{summary}, and \texttt{condition}. Making \edgereasoninfer~more general is a design choice backed by two reasons. First, the five labels do not fully cover the corner cases in \edgereasoninfer~edge. For instance, the third subtype of \edgereasoninfer~(Table \ref{tab:edge-labels}) annotates the inductive reasoning process that connects a single example to a general pattern; it does not directly correspond to \texttt{summary} (denotes many-to-one restatement) or \texttt{explanation} (examples are not a reason why the pattern is true). Second, dividing \edgereasoninfer~does not alter how to treat different labels in downstream tasks; i.e., the process for verifying the reasoning correctness of \texttt{summary} and \texttt{explanation} pairs will be identical.

However, for other reasoning-focused behaviors than inference, ReasoningFlow is significantly more expressive. First, while ReasoningFlow includes many long-distance relations that connect different stages of reasoning (\edgeplanbacktrack, \edgevalidatesupport, \edgevalidateattack), RST lacks these long-distance relations due to projectivity constraints. Second, while RST annotates attitude to the nucleus via a single \textit{evaluation} edge, ReasoningFlow also annotates the sentiment (\edgereflectpositive, \edgereflectuncertain, \edgereflectnegative). Finally, RST covers relations between a plan and its implementation via \texttt{Manner-Means} or \texttt{Enablement}, but does not cover general cases as shown in the annotation guide for \edgereasonexecute.



\paragraph{Penn Discourse TreeBank (PDTB)} Another popular branch of discourse parsing is PDTB \citep{prasad_penn_2008}, which annotated various discourse relations on top of the Penn TreeBank corpus of news articles from the Wall Street Journal. They annotate independent binary relations between clauses, unlike a fully connected tree in RST.

PDTB edges are defined using the semantics of \textit{connectives} like "but" (\texttt{contrast}) or "Therefore" (\texttt{cause}). However, ReasoningFlow's edge definitions also include content-based relations that cannot be fully captured using connectives. For instance, \edgeplanproceed~is purely annotated based on whether the semantic content of the \nodeplanning~step is motivated by the previous steps, and such a relation has no one-to-one correspondence with any connective in English.


\subsubsection{Argumentation structure mining}

In argumentative texts, argumentative components (claims, evidence) exhibit directed argumentative relations that indicate the degree of support of a component for another. Persuasive Essays (PE) corpus \citep{stab_parsing_2017} is the representative resource for argumentative structure mining. PE annotates spans as \textit{major claims}, \textit{minor claims}, and \textit{premises} along with \textit{support} and \textit{attack} relationships between these spans. While the exact label set varies within datasets, the general idea of identifying claims and premises with support and attack relations mostly remains consistent across argumentation mining literature \citep{lauscher_argument-annotated_2018, habernal_mining_2024}.

While argumentation mining frequently targets argumentative text like persuasive essays, academic papers, and legal judgments, a reasoning trace for objective tasks (e.g., math) can also be viewed as a type of argumentation. In this analogy, \texttt{major claims} corresponds to the final answer (\nodeconclusion), where \nodefact~and \nodereasoning~roughly correspond to leaf evidence (base facts) and minor claims derived from evidence, respectively. In terms of edges, \textit{support} relation in argumentation mining corresponds to \edgereasoninfer~and \edgevalidatesupport\footnote{Seldomly, \edgereflectpositive~edge (affirmation of previous nodes) if the destination \nodereflection~node includes a specific reason -- as in \textit{"I think that's it, because the boiling point of Benzo derivatives are typically around 180$^\circ$C to 330$^\circ$C."} -- instead of only having filler phrases like \textit{"I think that's it."}}.

Compared to argumentation mining, ReasoningFlow includes more fine-grained discourse relations for diverse reasoning behaviors. We claim that ReasoningFlow's expressive label sets possess the potential to expand the scope of argumentation mining. For instance, existing works seldom identify rare argumentation strategies like hypothetical reasoning (e.g., \textit{reductio ad absurdum}) or analogical reasoning \citep{walton_argumentation_2008}; ReasoningFlow's \nodeassumption~and \nodeexample~nodes naturally distinguishes these strategies from common deductive argumentation.

\subsubsection{Natural deduction}

Natural deduction is a calculus for formal logic that involves \textit{natural} inference rules that align with human instincts, such as \textit{modus ponens}, \textit{modus tollens}, and hypothetical syllogism.

Natural deduction systems can be classified by their reasoning directions: bottom-up and top-down. Bottom-up systems combine base facts to repeatedly deduce new conclusions until the conclusion is reached \citep{lifschitz_answer_2019, tafjord_proofwriter_2021, creswell_selection-inference_2023}. On the other hand, top-down systems start from the desired conclusion (goal) and recursively decompose it into subgoals until verified \citep{wielemaker_overview_2003, kazemi_lambada_2023, lee_symba_2025}. In ReasoningFlow, bottom-up deductive reasoning and top-down decomposition/planning are both captured via \edgereasoninfer~and \edgeplandecompose~edges, respectively.

Furthermore, assumptions play a significant role in natural deduction, which motivated the \nodeassumption~nodes in ReasoningFlow. In natural deduction, a \textit{subproof} is a nested block of reasoning where a statement is temporarily assumed, and the block is discharged when the expected conclusion is proved. It is frequently utilized by inference rules like conditional proofs (To prove $p\rightarrow q$, assume $p$ and show $q$), or-elimination (Given $p \lor q$, prove $p \rightarrow r$ and $q \rightarrow r$ to show $r$), and proof-by-contradiction (To prove $a$, show that $\neg a$ leads to a contradiction).


\subsubsection{Cognitive science}

Whether LLMs are \textit{meant} to resemble human cognitive behaviors during the Chain-of-thoughts reasoning is a debatable question \citep{bao_how_2025, chen_reasoning_2025, hao_training_2024}. However, it is widely accepted that cognitive science provides a useful guide for understanding LLM reasoning \citep{zhang_system_2026, liu_mind_2025}. In this section, we use \citet{kargupta_cognitive_2025}'s exhaustive taxonomy of cognitive concepts related to human and LLM reasoning \textit{operations}, and their relation to ReasoningFlow.

\paragraph{Reasoning Navigation.} 
The three core operations for navigating the reasoning process are \texttt{Forward chaining}, \texttt{Backward chaining}, and \texttt{Backtracking}. Forward chaining recursively combines known facts towards the final goal, while backward chaining decomposes goals into prerequisites. Similar to bottom-up and top-down inference in natural deduction, forward chaining and backward chaining directly correspond to \edgereasoninfer~and \edgeplandecompose. Finally, backtracking is the ability to revisit and correct prior reasoning paths, often implemented as a depth-first search that returns to the previous decision point to explore alternatives. In ReasoningFlow, backtracking is denoted by \edgeplanbacktrack~or \nodeassumption~connected via \edgeplanproceed, which directly connects the failed strategy/assumptions to a new alternative.

\paragraph{Verification.} \texttt{Verification} evaluates whether the reasoning steps are consistent, plausible, and coherent with the provided facts or the world state, which is captured by \edgeplanverify. While there are orthogonal criteria for evaluating the reasoning steps, such as coherence or utility \citep{lee_evaluating_2025}, we do not further distinguish the detailed intent because most self-verification processes in LRMs are focused on logical validity.

\paragraph{Modifying knowledge representations.} \citet{kargupta_cognitive_2025} also proposes a wide terminology for operations that derive new information from existing ones.

\texttt{Pattern recognition} involves identifying recurring \textit{templates}, or applying a general idea to a problem-specific knowledge. In reality, this is captured by determining which rule or idea to apply to the current pool of knowledge. It is captured by ReasoningFlow's \edgeplanproceed~that derives a general plan from previous steps.
\texttt{Abstraction} is when one detects abstract, generalizable patterns from specific instances, also known as inductive reasoning. In ReasoningFlow, this corresponds to when multiple \nodeexample~(that \edgereasonexemplify~a same concept) collectively deduce a single conclusion. ReasoningFlow does not distinguish between abstraction and ordinary deductive reasoning at the edge level because they can be identified by node labels.

\texttt{Representational restructuring} involves reformulating the goal to obtain new insights that lead to a better solution. \citet{kargupta_cognitive_2025} applies the widest possible definition of restructuring for analyzing reasoning traces in the focus of \textit{cognitive} behaviors, ranging from decomposing objective subgoals from a subjective problem (e.g., changing \textit{Which cellphone is better?} to \textit{Which cellphone has more memory?} to restructuring the given equation for a better solution (e.g., partial fraction decomposition). However, at the discourse structure level, we claim that these instances fundamentally differ in their semantics, e.g., explicitness or logical equivalence. Hence, we apply the most appropriate edge labels for each of the cases, e.g., \edgeplandecompose~for the former example and \edgereasoninfer~for the latter.

\section{Automatic annotation (\S\ref{sec:automatic-annotation}) details}
\label{sec:appendix-automatic-annotation}

\subsection{Implementation details}
\label{subsec:appendix-automatic-annotation-implementation-details}

\paragraph{Node segmentation.} When asking LLMs to segment an entire reasoning trace into nodes, we have frequently observed two failure modes: (1) segmented nodes do not exactly match the original trace, and (2) some segments have excessive length ($10{,}000+$ characters). To address these issues, we first apply rule-based segmentation and then use LLMs to further segment them into atomic units.

We begin the segmentation process by rule-based detection of paragraphs. Instead of using all double newlines \texttt{\textbackslash n\textbackslash n} as paragraph delimiters for reasoning traces, we split only when the preceding paragraph ends with a period, and the next paragraph starts with a capital letter. This is to ensure that a single semantic unit (e.g., equation series in a single \texttt{begin}...\texttt{end} block) is not mistakenly segmented into two parts. 

After the initial coarse segmentation, we leverage LLMs to further segment the chunks into ReasoningFlow nodes. If any segments are not perfectly aligned with the original string (often whitespaces or \TeX~symbols), we use dynamic programming to find the optimal alignment between the remaining segments, maximizing the total longest common subsequence length across all unmatched components.

\paragraph{Node classification.} Instead of classifying the nodes independently, we annotate all nodes simultaneously in a single inference, based on our preliminary results. We provide node definitions and examples for each subtype, as shown in Table \ref{tab:node-labels}.

\paragraph{Post-hoc annotation of Conclusion nodes.} Every reasoning trace must contain at least one \nodeconclusion~node by definition; however, we find that initial annotation occasionally omits one even if instructed. To prevent this, we adopt a two-stage procedure where the LLMs first classify all nodes with labels excluding \nodeconclusion, and then identify which should be labeled as \nodeconclusion. This design ensures that every reasoning trace contains at least one \nodeconclusion~node.

\paragraph{Edge detection and classification.} We consider three possible implementations for edge detection: (1) annotating all edges simultaneously, (2) annotating incoming edges per node, and (3) annotating all node pairs individually (\textit{dyadic}). Following \citet{mukherjee_premise-augmented_2025}, we adopt (2) for its balance of annotation quality and computational efficiency. For each node, we prompt LLMs to identify the minimal complete set of predecessors along with their corresponding edge labels.

ReasoningFlow edges are type-dependent: each node type admits only a restricted subset of incoming edge labels. For example, \edgereflectpositive~can only flow into \nodereflection~by definition. We therefore provide the set of permitted edge labels for each node type when querying for predecessors, reducing label ambiguity.

The prompts for LLM annotation in all stages are presented in Appendix \ref{sec:appendix-prompts}.

\subsection{Automatic annotation performance}
\label{subsec:appendix-model-performance-comparison}

\begin{table}[tb]
\footnotesize
\centering
\begin{tabular}{l|cc|cc}
\toprule
\multirow{2}{*}{\textbf{Model}}     & \multicolumn{2}{c|}{F1 score} & \multicolumn{2}{c}{Krippendorff's $\alpha$} \\
                   & NC     & EDC    & NC     & EDC    \\
\midrule
GPT-5-mini         & 0.792  & 0.425 & 0.741  & 0.444  \\
GPT-5.1            & 0.833  & 0.574 & 0.761  & 0.604  \\
\midrule
Gemini-2.5-Flash   & 0.811  & 0.535  & 0.749  & 0.576  \\
Gemini-3-Flash     & 0.865  & 0.583  & 0.820  & 0.607  \\
Gemini-3.1-Pro     & 0.859  & 0.646  & 0.812  & 0.677  \\
\bottomrule
\end{tabular}
\caption{F1 scores and inter-annotator agreement (Krippendorff's $\alpha$) measured between human annotators and LLM models with ground-truth segmentation. F1 scores are macro-averaged for each reasoning trace. Gemini-3-Flash achieves the best score in node classification (NC), while Gemini-3-Pro exhibits strong edge detection/classification capability (EDC).}
\label{tab:human-llm-annotator-agreement}
\end{table}

\begin{figure*}
    \centering
    \includegraphics[width=\linewidth]{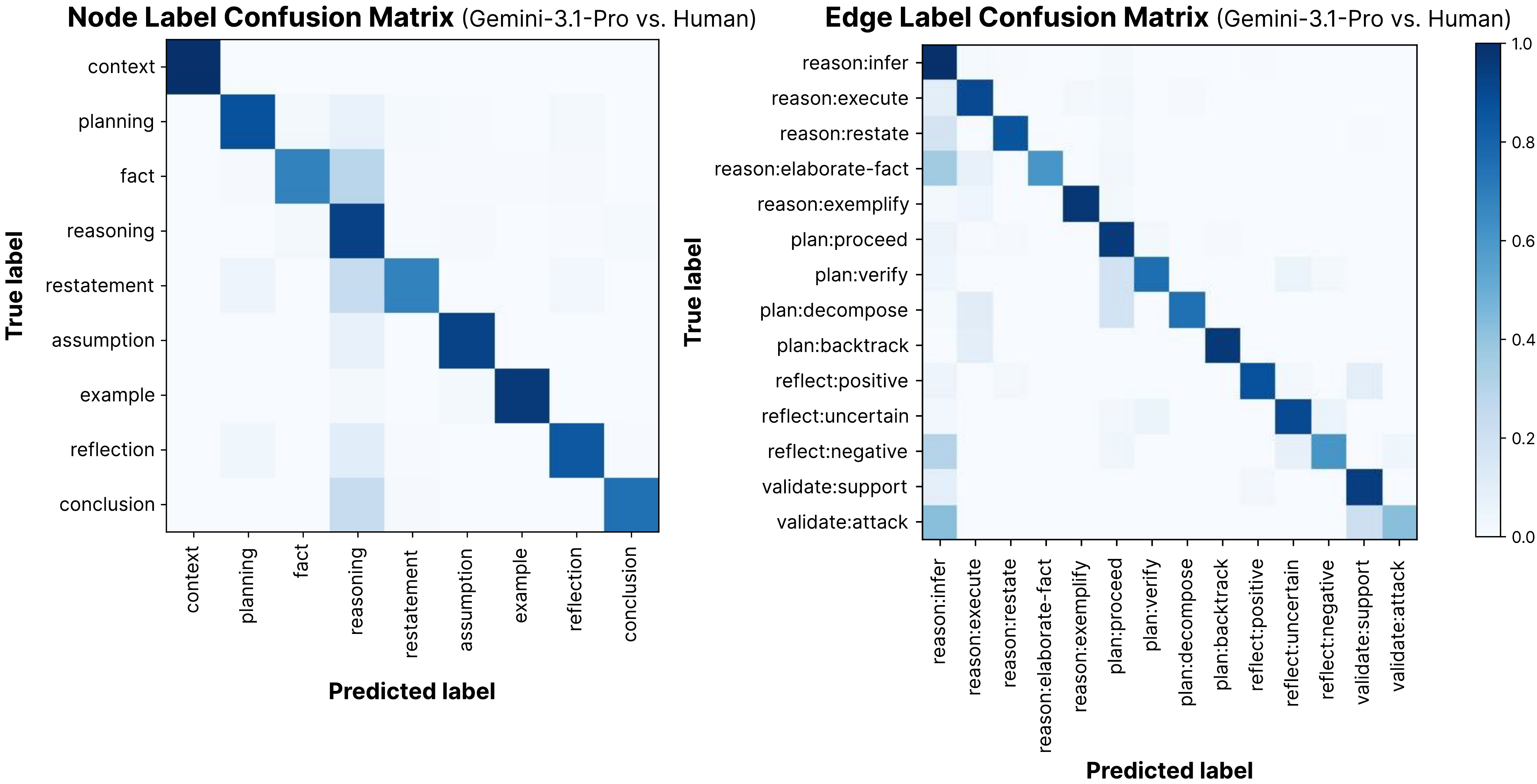}
    \caption{Confusion matrix (row-normalized) of Gemini-3.1-Pro annotations compared to human annotations.}
    \label{fig:confusion-matrix}
\end{figure*}

\paragraph{Model performance.} To select the LLM to use for automatic annotation of ReasoningFlow, we compare various LLMs using the manually annotated set in Section \ref{sec:manual-annotation}.

Table \ref{tab:human-llm-annotator-agreement} reports the F1 scores for node classification (NC) and edge detection/classification (EDC)\footnote{For EDC F1 score, we do not treat \texttt{no-edge} as a positive label, unlike the calculation of Krippendorff's $\alpha$.}, alongside inter-annotator agreement metrics for direct comparison with human-human agreement in Table \ref{tab:inter-annotator-agreement}. For Node Classification, Gemini-3-Flash demonstrates the best performance overall, while for Edge Detection/Classification, Gemini-3-Pro exhibits strong performance, being the only model to achieve $\alpha\geq0.67$ for EDC. All results are performed with a single run of greedy decoding.

Following common practices in discourse parsing literature \citep{marcu_rhetorical_2000, morey_how_2017}, we compare models using the same ground-truth segmentation as manual annotations. In the following section, we show that automatic segmentation generates different boundaries from human predictions, but the final discourse structures are topologically compatible in most cases.

\paragraph{Confusion matrix analysis.} Figure \ref{fig:confusion-matrix} shows the confusion matrix of Gemini-3.1-Pro. Some common mistakes in node classification include classifying non-\nodereasoning~nodes (\nodefact, \nodereflection, \nodeconclusion) as \nodereasoning. The edge confusion matrix shows that misclassification of node labels directly affects the edge annotations, as the model confuses \edgereasonelaboratefact (often  $\rightarrow$\nodefact), \edgereflectnegative ($\rightarrow$\nodereflection), and \edgevalidateattack ($\rightarrow$\nodeconclusion) with reasoning nodes.

\subsection{Automatic node segmentation}

\begin{figure*}
    \centering
    \includegraphics[width=\linewidth]{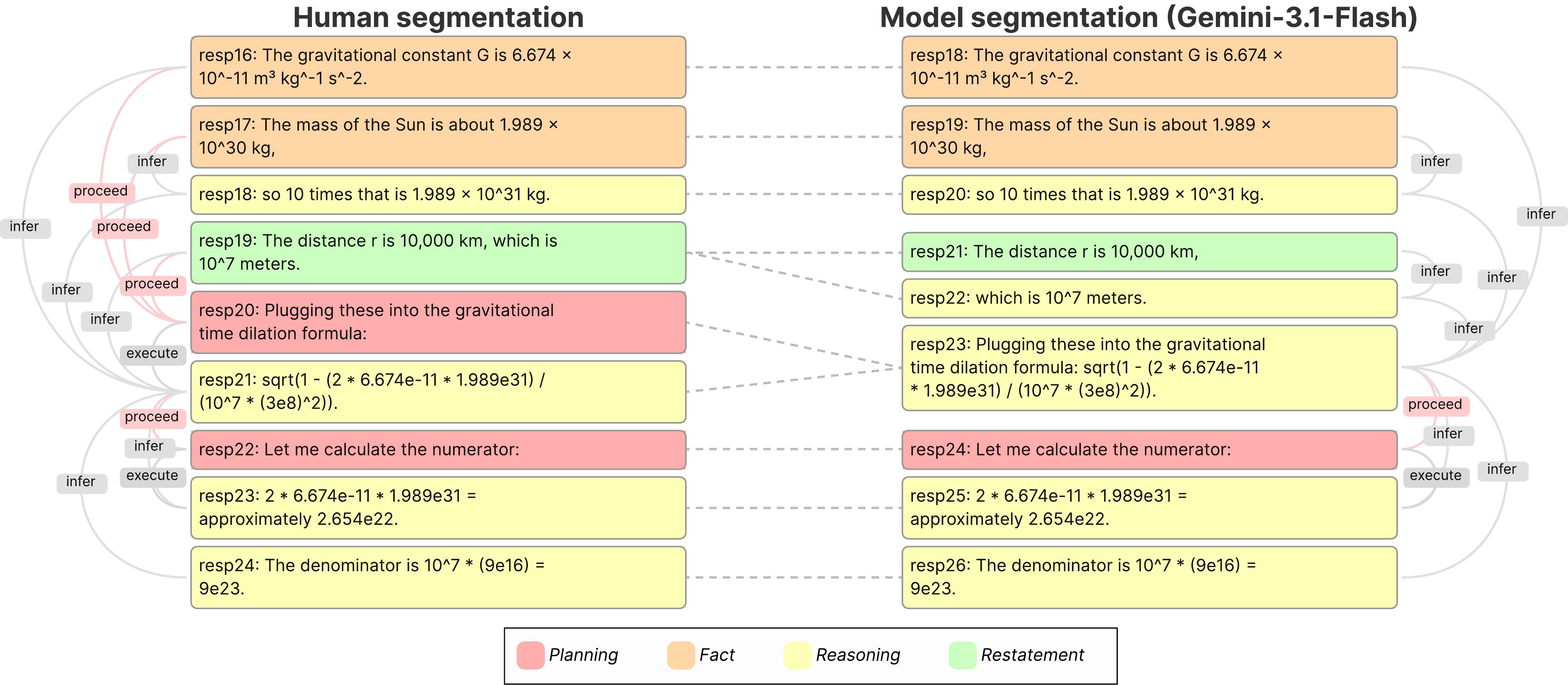}
    \caption{Example of a misalignment between human and LLM segmentation. Even if segmentations are misaligned, the overall structures remain semantically compatible. For instance, human segmentation's \texttt{resp21} node is topologically identical to model segmentation's \texttt{resp23} node, except \texttt{resp20$\rightarrow$resp21} that connects the first and second half of the model's \texttt{node23} node.}
    \label{fig:segmentation-diff-example}
\end{figure*}

In this section, we explore the effects of errors in automatic node segmentation. Due to the constraint that all nodes should be a sentence at longest, nearly all (95.6\%) misalignments in the test set are one-to-many misalignments, i.e., one side does not segment a sentence, while the other further segments one. Among these one-to-many misalignments, LLMs prefer more fine-grained segmentation than humans, as the number of over-segmentations (64.2\%) nearly doubles under-segmentations (35.8\%).

How do these segmentation misalignments affect the downstream ReasoningFlow structure? We qualitatively find that these misalignments between humans and LLMs do not cause significant effects in the overall structure. Figure \ref{fig:segmentation-diff-example} compares automatic ReasoningFlow annotation over human and LLM segmentations; while the segmentation is misaligned, the topology of the overall graph remains semantically consistent. This is similar to discourse parsing, where a failure to segment two sibling leaf spans casts minimal effect on the global tree structure \citep{nguyen_rst_2021}.

\subsection{Manual verification}
\label{subsec:appendix-manual-verification}

To ensure that the quality of automatically annotated ReasoningFlow data is acceptable, a single human annotator reviewed annotations for 30 randomly sampled reasoning traces (10 per dataset). Due to annotation resource constraints, we report \textit{acceptability judgments} by comparing LLM annotations with human corrections (rather than annotating from scratch), following prior work \citep{mukherjee_premise-augmented_2025, zeng_rejump_2025}. Specifically, a human manually edited the LLM-produced annotation, updating the node/edge labels and adding/removing edges. Humans are instructed not to modify the segmentation.

The results indicate that LLM-based annotations are highly agreeable, achieving an F1 score of 0.909 for node classification (NC) and 0.869 for edge detection/classification (EDC). Note that evaluating with from-scratch human annotation instead of acceptability (Table \ref{tab:human-llm-annotator-agreement}) will likely reduce the inter-annotator agreement scores.

\subsection{Computation cost}

We annotate the ReasoningFlow dataset with the best LLM available (Gemini-3.1-Pro and Gemini-3-Flash) to achieve annotation accuracy. In the first pass, we annotate all traces with Gemini-3-Pro for edge detection/classification and Gemini-3-Flash for the remaining components. However, 289 of the first 1,260 annotations failed three times in a row due to the instability of the Gemini-3.1-Pro API. For these instances, we fell back to the second pass, only using Gemini-3-Flash for all annotation tasks.

At the time of the experiment, Gemini-3-Pro API cost \$2.00/1M input tokens and \$12.00/1M output tokens; Gemini-3-Flash cost \$0.50/1M input tokens and \$3.00/1M output tokens. The first pass (Pro and Flash) cost a total of \$1,032.98 with 442M input tokens and 12.3M output tokens (both models combined), and the second pass (Flash only) cost \$1,571.59 with 754M input tokens and 5.2M output tokens.


\section{Node quality annotation (\S\ref{subsec:node-quality}) details}
\label{sec:appendix-node-quality}

\subsection{Stepwise evaluation with PARC}

To evaluate the validity (logical correctness) of each node, we employ the LLM-as-a-judge approach \citep{gu_survey_2024}. We prompt the LLM with a step and its context (question, previous steps), and make it reason about whether the given step is logically correct or incorrect based on the context. As ReasoningFlow includes extremely long reasoning traces, it is not feasible to provide \textit{all} previous steps as the evaluation context. Instead, we prune the context using the ReasoningFlow graph by selecting all preceding nodes within a graph distance of 2. As explored in non-reasoning models, context pruning improves verification performance by removing irrelevant context, while reducing the inference cost \citep{ling_deductive_2023, mukherjee_premise-augmented_2025, lee_evaluating_2025}.

Another benefit of using graph structures for stepwise evaluation is the distinction between direct and propagated errors. \citet{mukherjee_premise-augmented_2025} defines direct errors as steps that are erroneous but their premises are all correct, and propagated errors as steps that are correctly inferred from erroneous previous steps. As it is challenging to decide whether the propagated error should be considered an error \citep{lightman_lets_2024, zheng_processbench_2025}, the ternary labels (correct, direct error, propagated error) significantly reduce ambiguity.

\begin{table}[t]
\centering
\footnotesize
\begin{tabular}{lr}
\toprule
\textbf{True error?} & \textbf{Count} \\
\midrule
Yes      & 94 (61.4\%) \\
Somewhat & 29 (19.0\%) \\
No       & 30 (19.6\%) \\
\midrule
Total & 153 (100\%) \\
\bottomrule
\end{tabular}
\caption{Manual verification of PARC error detection accuracy, tested for 50 files with PARC-detected errors.}
\label{tab:parc-performance}
\end{table}

We also perform manual verification of PARC results, judging whether the PARC-predicted errors are true errors (i.e., \textit{precision}). Table \ref{tab:parc-performance} shows that most of the errors (80.4\%) are clearly or somewhat errors, indicating that the combination of PARC and ReasoningFlow is highly effective when detecting errors in LRM traces. While not directly comparable, the precision is significantly higher than the reported precision in DeltaBench \citep{he_can_2025} for detecting errors in LRM traces (GPT-4-Turbo 37.4\%).

\paragraph{Error detection with PRMs.} Process Reward Models (PRMs) \citep{uesato_solving_2022, lightman_lets_2024} are LLM-based classifiers specifically trained to predict whether the given step is correct or not. However, we do not apply PRMs for several reasons. First, state-of-the-art PRMs like Qwen2.5-Math-PRM-72B \citep{zhang_lessons_2025} are not trained to evaluate errors in long LRM traces. Consequently, 24-51\% of their error predictions in ReasoningFlow are \nodeplanning~or \nodereflection~, which cannot be erroneous by definition, indicating that PRMs are not directly applicable for evaluating LRM traces. Furthermore, PRMs often have limited context length (e.g., Qwen2.5-Math-PRM can take 4096 tokens), making it infeasible for detecting errors beyond that limit. While recent works explore PRMs specifically trained for reasoning models \citep{zou_reasonflux-prm_2025, xie_towards_2026}, they have not been verified against LRM-focused stepwise evaluation benchmarks (e.g., \citet{he_can_2025}) or have not released the PRM checkpoint.

\subsection{Argument quality scoring with AQR}

\begin{table}[t]
\centering
\footnotesize
\begin{tabular}{lr}
\toprule
\textbf{Arg. Match?} & \textbf{Count} \\
\midrule
Yes      & 338 (84.3\%) \\
Somewhat & 52 (13.0\%) \\
No       & 11 (2.7\%) \\
\midrule
Total & 401 (100\%) \\
\bottomrule
\end{tabular}
\caption{Manual verification of argument alignment between ReasoningFlow's Reasoning nodes and AQR-30k's human-annotated arguments on the ArgKP subset.}
\label{tab:argumentation-align-performance}
\end{table}

To estimate the quality of reasoning nodes in the ArgKP dataset \citep{bar-haim_arguments_2020}, we use AQR-30k \citep{gretz_large-scale_2020}. AQR-30k includes crowdsourced arguments with their scores for 71 debate topics, which include all 24 topics in ArgKP. They propose WA (Weighted Average) scores to reduce the contributions of unreliable score annotators, which we use as the argument score.

We observed that AQR-30k includes logically equivalent arguments with significantly different scores. For instance, the two arguments against capital punishment: \textit{"Capital punishment is against god's will."} and \textit{"Only god decides who lives and who dies; capital punishment should not exist."} are almost equivalent, but the WA-scores assigned to them are 0.42 and 1.0, respectively. To reduce variance, we prompt Gemini-3-Flash to map a ReasoningFlow node to multiple AQR-30k arguments that are logically equivalent, and take the average score of the mapped AQR-30k arguments as the node quality. Table \ref{tab:argumentation-align-performance} shows the manual verification results of the matching between randomly sampled 50 nodes and 401 arguments, proving that such LLM-based matching is highly precise.

\section{ReasoningFlow statistics (\S\ref{sec:statistics}) details}
\label{sec:appendix-stats}

\begin{figure*}
    \centering
    \includegraphics[width=\linewidth]{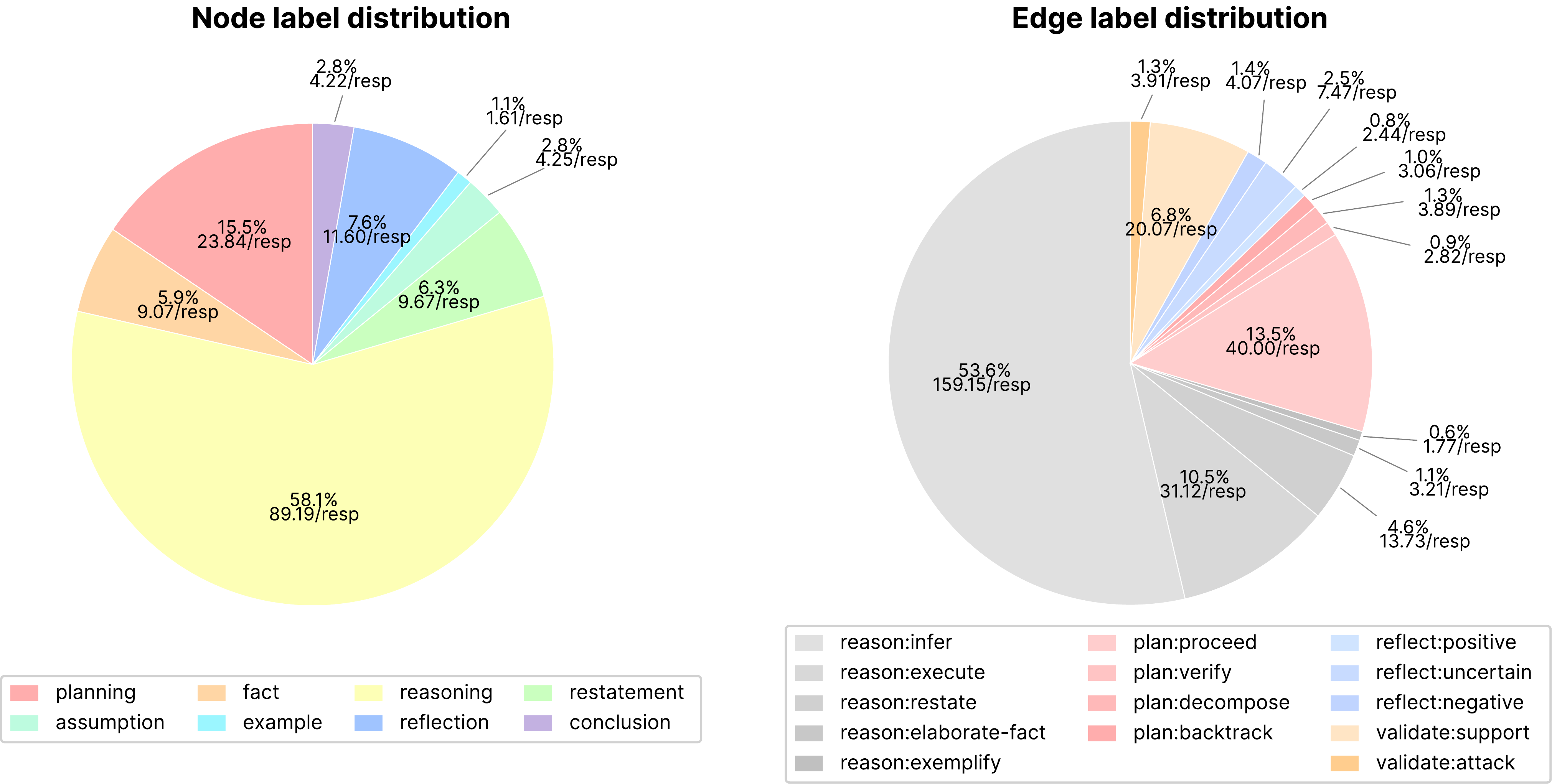}
    \caption{Node and edge label distribution in ReasoningFlow. \texttt{/resp} indicates how many times a node/edge occurs in a single LLM response.}
    \label{fig:label-distribution}
\end{figure*}

\begin{table}[t]
\centering
\footnotesize
\begin{tabular}{p{5.5cm}p{1.2cm}}
\toprule
\textbf{Triplet} & \textbf{Loading} \\
\midrule
\multicolumn{2}{l}{\textbf{PC1 (41.0\%)}} \\
\midrule
\nodecontext--\edgereasoninfer$\rightarrow$\nodereasoning & -0.321  \\
\nodeplanning--\edgeplanproceed$\rightarrow$\nodeplanning & +0.311  \\
\nodereasoning--\edgeplanproceed$\rightarrow$\nodeplanning & -0.277  \\
\midrule
\multicolumn{2}{l}{\textbf{PC2 (20.2\%)}} \\
\midrule
\nodecontext--\edgereasoninfer$\rightarrow$\nodereasoning & +0.395  \\
\nodeconclusion--\edgevalidatesupport$\rightarrow$\nodeconclusion & +0.377  \\
\nodereasoning--\edgereasoninfer$\rightarrow$\nodereasoning & -0.364  \\
\bottomrule
\end{tabular}
\caption{Top-3 PCA loadings for first and second principal components in triplet analysis (Figure \ref{fig:triplet-distribution}(a)).}
\label{tab:pca_loadings}
\end{table}

\paragraph{Label distributions.} Figure \ref{fig:label-distribution} shows the node and edge label distribution in ReasoningFlow. Since most tasks are reasoning-focused, \nodereasoning~occupies the largest portion of the nodes. While \nodeassumption~and \nodeexample~are rare (2.8\% and 1.1\%, respectively), they carry unique semantics for identifying reasoning behaviors like assumption scoping and inductive reasoning and cannot be neglected. For the edges, \edgereasoninfer~is the most common label as it is the primary label for connecting the most common \nodereasoning~nodes.

\paragraph{Triplet analysis details}
To faithfully analyze triplet distribution using PCA, we use Hellinger PCA, i.e., PCA on vectors of $\sqrt{\text{probability}}$ \citep{lebret_word_2014}. Taking the square root of probability improves fairness in distance for sparse probabilities. For instance, distance metric $| t_1 - t_2 |$ treats the difference $.5-.51$ identical to $.01-.02$, while $| \sqrt{t_1} - \sqrt{t_2} |$ assigns more weight to the latter. Since ReasoningFlow has a very sparse triplet distribution, we find Hellinger PCA a more appropriate approach than PCA on the raw probability vectors.

Regarding Figure \ref{fig:triplet-distribution}(a)'s two principal components (PC), the top 3 triplets with the largest loadings (i.e., $v_{\{1,2\}}^\top e_i$ are presented in Table \ref{tab:pca_loadings}. Compared to AIME and GPQA (\texttt{-PC1}), ArgKP traces (\texttt{+PC1}) are characterized by chained \nodeplanning~nodes that enumerate minor claims (\textit{"First, evidence supporting flag burning should be allowed due to freedom of speech: (evidence) Second, ..."}), leading to frequent \edgeplanproceed~edges. GPQA traces (\texttt{+PC2}) frequently reference \nodecontext~nodes that include important information like conditions of a reaction, and global verification (\nodeconclusion--\edgevalidatesupport$\rightarrow$\nodeconclusion) generates uniform answers compared to more fluctuating final answers in AIME (\texttt{+PC2}).

\section{Reasoning behaviors (\S\ref{sec:behaviors}) details}
\label{sec:appendix-behaviors}

\begin{table}[t]
\centering
\footnotesize
\begin{tabular}{ll}
\toprule
\textbf{Local Verification} (\S\ref{subsec:local-verification}) & \\
\texttt{(Reasoning)}--\edgevalidatesupport$\rightarrow$\nodereasoning & \\
\texttt{(Reasoning)}--\edgevalidateattack$\rightarrow$\nodereasoning & \\
\texttt{(Reasoning)}--\edgeplanverify$\rightarrow$\nodeplanning & \\
\midrule
\textbf{Self-Reflection} (\S\ref{subsec:self-reflection}) \\
\texttt{(Reasoning)}--\edgereflectpositive$\rightarrow$\nodereflection & \\
\texttt{(Reasoning)}--\edgereflectuncertain$\rightarrow$\nodereflection & \\
\texttt{(Reasoning)}--\edgereflectnegative$\rightarrow$\nodereflection & \\
\midrule
\textbf{Assumption} (\S\ref{subsec:assumption}) & \\
\textit{Proof-by-contradiction} & \\
\nodeassumption--\edgevalidateattack$\rightarrow$\nodereasoning & \\
\textit{Switch-case}\\
\nodeassumption--\edgeplanproceed$\rightarrow$\nodeassumption & \\
\nodeassumption--\edgeplanbacktrack$\rightarrow$\nodeassumption & \\
\bottomrule
\end{tabular}
\caption{Subgraph patterns for reasoning behavior analyzed in Section \ref{sec:behaviors}. For local verification and self-reflection analysis on reasoning datasets (AIME, GPQA), \texttt{(Reasoning)} includes \nodefact, \nodereasoning, \noderestatement\, and \nodeconclusion.}
\label{tab:behavior-analysis-patterns}
\end{table}

\paragraph{Triplet pattern details.} Table \ref{tab:behavior-analysis-patterns} shows the triplet patterns used for identifying behaviors in Section \ref{sec:behaviors}.

ReasoningFlow allows analyzing more complex structures beyond triplets, i.e., a subgraph spanning from the \nodeconclusion~node to retrieve \textit{all} premises that support the conclusion, or the tree of \nodeplanning~nodes connected via \edgeplandecompose~and\edgeplanproceed~to analyze the top-down planning ability. We leave analyses of larger structures beyond triplets as future work.

\paragraph{Robustness against annotator models.} Are the reasoning behavior analysis results robust to the selection of the annotation model? We compare the self-reflection results (Section \ref{subsec:self-reflection} on DeepSeek-R1's AIME traces, with GPT-5.1 as the annotator model for nodes and edges.

Table \ref{tab:self_reflection_annotator_compare} shows that while the actual numbers differ, the trend is preserved: positively reflected steps are most likely to be correct, and negatively reflected steps are most likely to be erroneous.

\begin{table}[t]
\centering
\footnotesize
\setlength{\tabcolsep}{6pt}
\begin{tabular}{lrrr}
\toprule
\textbf{Annotator} & \textbf{Corr.} & \textbf{Prop.} & \textbf{Error} \\
\midrule
\multirow{3}{*}{Gemini-3}
  & 78.4 & 20.0 & 2.4 \\
  & 59.6 & 29.8 & 10.7 \\
  & 49.1 & 33.8 & 17.1 \\
\midrule
\multirow{3}{*}{GPT-5.1}
  & 78.6 & 17.9 & 3.6 \\
  & 73.5 & 19.5 & 7.0 \\
  & 63.2 & 16.3 & 22.4 \\
\bottomrule
\end{tabular}
\caption{Comparison between Gemini-3 family vs. GPT-5.1 annotations in analyzing self-reflection and step-level correctness, using DeepSeek-R1 traces from AIME dataset. Corr. and Prop. stands for Correct and Propagated, respectively.}
\label{tab:self_reflection_annotator_compare}
\end{table}

\section{Stepwise evaluation (\S\ref{sec:validity-evaluation}) details}
\label{sec:appendix-validity-evaluation}

\begin{table*}[h]
\centering
\footnotesize
\begin{tabular}{llrrrr}
\toprule
Domain & Model & Total & \textit{Unused} & \textit{Neglected} & \textit{Faithful} \\
\midrule
AIME & QwQ-32B              & 123  & 98 (79.7\%)  & 12 \hphantom{0}(9.8\%)  & 13 (10.6\%) \\
AIME & DeepSeek-R1          & 215  & 192 (89.3\%) & \hphantom{0}8 \hphantom{0}(3.7\%)  & 15 \hphantom{0}(7.0\%) \\
AIME & GPT-oss         & \hphantom{0}61   & 39 (63.9\%)  & \hphantom{0}2 \hphantom{0}(3.3\%)  & 20 (32.8\%) \\
AIME & Qwen2.5-32B & \hphantom{0}83   & 28 (33.7\%)  & \hphantom{0}3 \hphantom{0}(3.6\%)  & 52 (62.7\%) \\
AIME & DeepSeek-V3          & 200  & 180 (90.0\%) & \hphantom{0}0 \hphantom{0}(0.0\%)  & 20 (10.0\%) \\
\midrule
GPQA & QwQ-32B              & 3398 & 2813 (82.8\%) & 203 \hphantom{0}(6.0\%) & 382 (11.2\%) \\
GPQA & DeepSeek-R1          & 2928 & 2529 (86.4\%) & 156 \hphantom{0}(5.3\%) & 243 \hphantom{0}(8.3\%) \\
GPQA & GPT-oss         & 1409 & 1060 (75.2\%) & 108 \hphantom{0}(7.7\%) & 241 (17.1\%) \\
GPQA & Qwen2.5-32B & 438  & 171 (39.0\%)  & 102 (23.3\%) & 165 (37.7\%) \\
GPQA & DeepSeek-V3          & 405  & 117 (28.9\%)  & 47 (11.6\%)  & 241 (59.5\%) \\
\bottomrule
\end{tabular}
\caption{Evaluating whether the erroneous nodes are used as a premise for the final answer. Following the taxonomy of Figure \ref{fig:validity-eval-examples}, three columns correspond to \textit{not reaching any conclusion} (\textit{Unused}), \textit{reaching a correct answer} (\textit{Neglected}), and \textit{reaching an incorrect answer} (\textit{Faithful}).}
\label{tab:validity-evaluation-full}
\end{table*}

\paragraph{Full results.} As explained in Section \ref{sec:validity-evaluation}, unused errors can be attributed to excessive backtracking that leads to abandoning incorrect paths without generating a valid conclusion, and neglected errors often arise from unfaithfully ignoring orthogonal errors. 

Table \ref{tab:validity-evaluation-full} shows the erroneous step distributions for each (model, dataset) configuration. As discussed in Section \ref{sec:validity-evaluation}, QwQ and R1 both generally exhibit both a high unused rate (>80\%) and a neglect rate ($\text{neglected}/(\text{neglected}+\text{faithful})$>30\%) for both datasets. This highlights that step-level errors seldom have an effect on final answers in these models. Therefore, solely relying on step-level validity evaluators for predicting final answer correctness of LRM traces, e.g., in Best-of-N decoding \citep{zou_reasonflux-prm_2025}, is likely to underperform.

GPT-oss and Non-reasoning models (Qwen2.5-32B, DS-V3) demonstrate low unused rates in AIME, while showing significantly higher rates in GPQA. We attribute this to the multiple-choice format of GPQA; even if the intermediate step contained errors, e.g., writing an incorrect SMILES representation of a molecule, it has a much higher chance of choosing the correct answer than in AIME due to the constrained final answer space. This finding extends previous claims that LLM reasoning can be unfaithful when exposed to answer leaks or unanswerable questions \citep{lanham_measuring_2023, balepur_which_2025} with quantitative evidence in reasoning trace structures.

\paragraph{Manual verification of unused errors.} To validate the findings in Section \ref{sec:validity-evaluation}, authors manually inspected 30 randomly sampled “Abandoned” nodes across LRMs and AIME/GPQA (15 each), checking whether missing ground-truth edges would have classified unused nodes otherwise. Only 10\% of the cases were due to the annotation missing an edge; 90\% of the unused cases were genuinely related to backtracking/local verification/assumption scopes.

In detail, the erroneous nodes were unused due to backtracking (trying a different strategy, 40.0\%), local verification (explicitly reflecting on and correcting previous steps, 33.3\%), and underlying assumptions being rejected (10\%), highlighting the importance of understanding reasoning behaviors for evaluating the effects erroneous steps have on the model's final answer. The remaining 6.7\% contained facts that are irrelevant and unnecessary for solving the problem, therefore not being connected to any subsequent nodes.

\paragraph{Robustness against annotator models.} To show that the result is consistent across different annotator models, we compare the analysis results on DeepSeek-R1's AIME traces, with GPT-5.1 as the annotator model for nodes and edges.

The rate of unused edges is nearly identical in both annotator models, marking 89.3\% and 89.7\% in Gemini and GPT, respectively. Therefore, the core finding "most of the errors are not used for deriving the final answer" remains the same regardless of the annotator models. However, the rate of neglected edges changes from 3.7\% to 9.2\%, and faithful edges from 7.0\% to 1.1\%, respectively. Qualitatively, this can be partially explained by GPT-5.1's lower edge-detection recall compared to Gemini models (Table \ref{tab:human-llm-annotator-agreement}).

\section{Mechanistic Interpretability (\S\ref{sec:thought-anchors}) details}
\label{sec:appendix-thought-anchors}

\begin{figure}
    \centering
    \includegraphics[width=\linewidth]{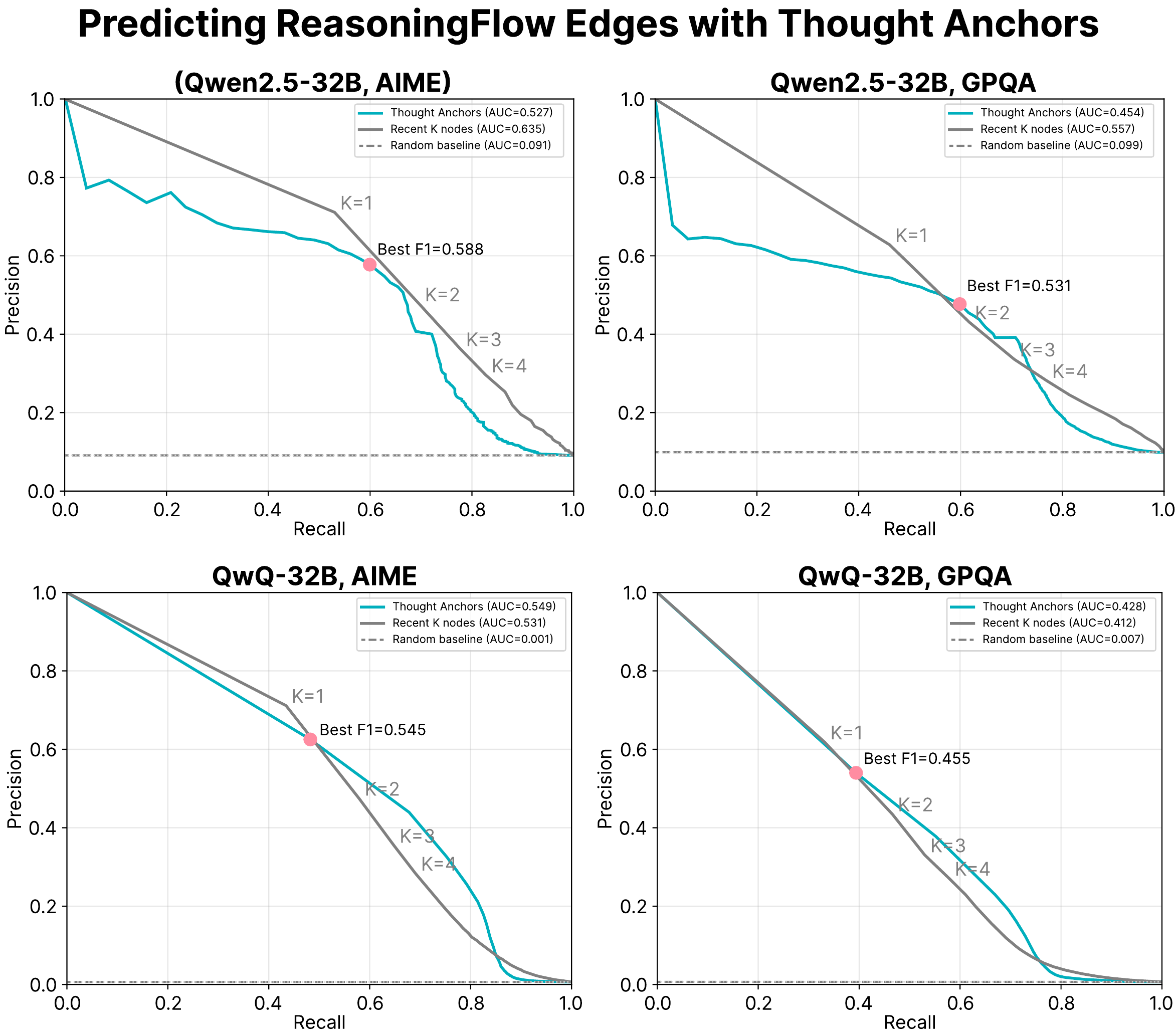}
    \caption{P-R curve of using Thought Anchors' scores to predict ReasoningFlow edges, showing that Thought Anchors are not more predictive than simply choosing $K$ most recent nodes in all four configurations.}
    \label{fig:thought-anchors-full}
\end{figure}

\paragraph{Full results.} For Section \ref{sec:thought-anchors}, we replicate the Thought Anchor's causal dependency analyses for Qwen2.5-32B and QwQ-32B, as larger models could not be loaded on the available device. Figure \ref{fig:thought-anchors-full} shows the result for both models for AIME and GPQA. The results indicate that Thought Anchor's unsupervised detection of causal dependencies is not significantly better than selecting the closest $K$ nodes, showing the fundamental misalignment between semantic layers and causal dependencies.

Instead of causal masking, one can also analyze the \textit{hidden state representations} with supervised or unsupervised approaches (e.g., probing, sparse auto-encoders). For instance, \citet{zhong_chains_2026} shows that supervised probing on the difference of hidden state vectors $d_i-d_j$ can reconstruct ground-truth \edgereasoninfer~edges defined in synthetic reasoning benchmarks \citep{tafjord_proofwriter_2021}. 

\paragraph{Manual verification of mismatched edges.} To validate the findings in Section \ref{sec:thought-anchors}, we manually check 30 of the top 0.01\%-scoring Thought Anchor edges that are not present in ReasoningFlow. Among these mismatched edges, only 23.3\% are due to ReasoningFlow annotation errors; 76.7\% are either clearly not connected or only distantly connected (path length > 1) in ReasoningFlow. This reinforces our finding that mechanistic-level and discourse-level step dependencies are different, rather than an artifact of annotation errors in ReasoningFlow.

\section{Prompts}
\label{sec:appendix-prompts}

This section includes the prompts used for LLM-based automatic annotators. All prompts require JSON-formatted output. For \textit{Prompt 2: Node Classification}, we omit the definition of \nodeconclusion~, as they will be annotated using Prompt 3. 

\section{License statement}

The license status of all models and datasets is presented below.

\begin{itemize}
  \item \textbf{DeepSeek-V3}: DeepSeek License v1.0
  \item \textbf{DeepSeek-R1}: MIT
  \item \textbf{Qwen2.5-32B-Instruct}: Apache 2.0
  \item \textbf{QwQ-32B}: Apache 2.0
  \item \textbf{GPT-oss-120b}: Apache 2.0

  \item \textbf{STILL-2}: Apache 2.0
  \item \textbf{NuminaMath}: Apache 2.0
  \item \textbf{AIME 2024}: Apache 2.0 (community re-releases)
  \item \textbf{GPQA Diamond}: CC-BY 4.0
  \item \textbf{ArgKP}: MIT
\end{itemize}

\newpage


\begin{tcolorbox}[promptbox, title={\textcolor{white}{Prompt 1: Node Segmentation}}]
\small
Please split the following text into ``reasoning units''. Reasoning units are mostly sentences, but can change by whether the sentence includes complex reasoning intents.
These sentences need not start with an alphanumeric character.

\medskip\noindent You need to \textbf{further split} in these cases where the sentence can be segmented into parts with different semantic roles:
\begin{itemize}[leftmargin=*]
  \item A sentence separated by a colon, where the two parts are clearly planning and its implementation. If the planning is a single word, do not separate.\\
  \textit{Example:} ``Let's now find the value of a by using the equation 3a-4=7:'',\quad ``a=(7+4)/3=11/3.''

  \item A sentence that combines an externally verifiable fact and reasoning based on the fact. \textit{Note:} If the fact is not an absolute fact but dependent on the context/previous steps, do not split.\\
  \textit{Example:} ``Since the speed of light is 299,752,258m/s,'',\quad ``the kinetic energy of a particle would be approximately 1.5M Joules.''

  \item A sentence that links a reasoning-like phrase and a self-evaluation by punctuation (comma, colon).\\
  \textit{Example:} ``The final answer for this question is 5,'',\quad ``which seems plausible to me.''

  \item A sentence that links an assumption and its consequence.\\
  \textit{Example:} ``Assuming a=1,'',\quad ``the polynomial becomes $x^2 + bx + 2b - 1 = 0$.''
\end{itemize}

\noindent\textbf{Note:} Do not split where there are no connecting punctuations (semicolon, colon, comma, etc.)

\medskip\noindent\texttt{<<input>>}
\end{tcolorbox}

\begin{tcolorbox}[promptbox, title={\textcolor{white}{Prompt 1: Node Segmentation (Cont'd)}}]
\small

\medskip\noindent However, do \textbf{not} split:
\begin{itemize}[leftmargin=*]
  \item Simple headers or transition words.\\
  \textit{Example:} ``1. Computing the velocity of the spacecraft'';\quad ``Case 1: x=5'';\quad ``Wait, let's verify the previous step.''

  \item When the premise follows the conclusion in a single sentence.\\
  \textit{Example:} ``Cigarrettes should not be banned, since it provides tax income that can support the federal government.'';\quad ``This calculation seems off, because I know that the C-H bond length is around 1.06 to 1.10 \AA.'';\quad ``Since a and b are coprimes as proved above, we can ensure that there exists an inverse element of a for multiplication modulo b and vice versa.''

  \item When a colon denotes a simple fact and its description.\\
  \textit{Example:} ``The derivative of $x\cos(x)$ is: $\cos(x) - x\sin(x)$.''

  \item A single \LaTeX{} equation block.\\
  \textit{Example:} ``$[x^2 - 4y = 0 \;\backslash\backslash\; y = x^2/4]$'';\quad ``\texttt{\textbackslash begin\{aligned\}...\textbackslash end\{aligned\}}''

  \item An equation followed by a step explaining the symbols used.\\
  \textit{Example:} ``The kinetic energy $E = mv^2/2$, where $m$ is the mass and $v$ is the velocity.''
\end{itemize}

\medskip\noindent Return a JSON array of string sentences: \texttt{\{"units": ["..."]\}}. Preserve all punctuations and newlines exactly as given.

\medskip\noindent\texttt{<<input>>}
\end{tcolorbox}

\begin{tcolorbox}[promptbox, title={\textcolor{white}{Prompt 2: Node Classification}}]
\small
Reasoning traces are already segmented into nodes, which are syntactically non-overlapping segments ranging from clause to one sentence (exceptionally multi-line equations), and are semantically atomic.
Your job is to assign labels to all nodes according to the provided description and examples below. Make sure you do not miss any nodes.
Respond by a list of JSON Dicts: \texttt{\{"responses": [\{"node\_id": "(node id)", "label": "(label)"\}, ...]\}}

\medskip\noindent\textbf{Labels}

\medskip\noindent\texttt{<<node\_definitions>>}

\medskip\noindent\textbf{\underline{Classification Rules}}

Reasoning traces are already segmented into nodes. Your job is to assign labels to \textit{all} nodes. Make sure you do not miss any nodes.
Respond by a list of JSON Dicts: \texttt{\{"responses": [\{"node\_id": "(node id)", "label": "(label)"\}, ...]\}}

\medskip\noindent\texttt{<<input>>}
\end{tcolorbox}

\begin{tcolorbox}[promptbox, title={\textcolor{white}{Prompt 3: Conclusion Node Update}}]
\small
\textbf{Conclusion Node Update Guidelines}

\medskip\noindent\textbf{Conclusion.}\quad
This node includes the model's final answer for the question.
Conclusion nodes can directly indicate the final answer by ...

\textit{... (examples) ...}

\medskip\noindent Annotate which of the nodes should be changed to \texttt{conclusion} nodes.
Respond in JSON Dict: \texttt{\{"conclusion\_node\_ids": ["(node id)", "(node id)", ...]\}}.

\medskip\noindent\texttt{<<input>>}
\end{tcolorbox}

\begin{tcolorbox}[promptbox, title={\textcolor{white}{Prompt 4: Edge Labeling}}]
\small
Given a destination node in a reasoning trace, find the relevant source node and edge labels according to the provided taxonomy.
\begin{itemize}[leftmargin=*]
  \item \textbf{Source nodes} should contain necessary information for the destination's reasoning process. Source nodes should cover \textbf{all} crucial antecedents, logical premises, background knowledge, and core plans/ideas of the destination step.
  \item Note that in most cases, with most common exceptions in \texttt{reason:restate} and \texttt{validate:support/refute}, source nodes are placed closely within the destination node.
  \item Each edge label has information about ``possible source labels''. The selected source node and edge labels must match.
  \item For certain types of nodes (e.g., reflection, fact), there might be no source node. If no source node satisfies the definitions below, you can return an empty list.
\end{itemize}

\medskip\noindent\textbf{Edge label definitions}

\noindent\texttt{<<edge\_definitions>>}

\medskip\noindent Respond in JSON format by listing all edges connecting to the ``destination step'':
\begin{quote}\small
\texttt{[\{ "source\_node\_id": "(node id)", "label": "(label)" \}]}
\end{quote}

\medskip\noindent\textbf{Few-shot examples}

\noindent\texttt{<<few\_shot\_examples>>}

\medskip\noindent\textbf{Actual input}

\noindent\texttt{<<input>>}
\end{tcolorbox}

\end{document}